\documentclass{CUP-JNL-DTM}%

\usepackage{graphicx}
\usepackage{multicol,multirow}
\usepackage{amsmath,amssymb,amsfonts}
\usepackage{mathrsfs}
\usepackage{amsthm}
\usepackage{rotating}
\usepackage{appendix}
\usepackage[numbers,sort]{natbib}
\usepackage{ifpdf}
\usepackage[T1]{fontenc}
\usepackage{newtxtext}
\usepackage{newtxmath}
\usepackage{textcomp}
\usepackage{lipsum}
\usepackage[colorlinks,allcolors=blue]{hyperref}
\usepackage{multirow}
\usepackage[table,xcdraw]{xcolor}

\theoremstyle{definition}

\numberwithin{equation}{section}

\jname{}
\articletype{}
\begin{document}

\begin{Frontmatter}

\title[Article Title]{Knowledge-guided Machine Learning: Current Trends and Future Prospects}

% There is no need to include ORCID IDs in your .pdf; this information is captured by the submission portal when a manuscript is submitted. 
\author[1]{Anuj Karpatne}
\author[2]{Xiaowei Jia}
\author[3]{Vipin Kumar}

% \authormark{Author Name1 \textit{et al}.}

\address[1]{\orgdiv{Department of Computer Science}, \orgname{Virginia Tech}, \email{karpatne@vt.edu}}

\address[2]{\orgdiv{Department of Computer Science}, \orgname{University of Pittsburgh}, \email{xiaowei@pitt.edu}}

\address[3]{\orgdiv{Department of Computer Science and Engineering}, \orgname{University of Minnesota}, \email{kumar001@umn.edu}}

% \address[2]{\orgdiv{Division}, \orgname{Organization}, \orgaddress{\city{City}, \postcode{Pincode}, \state{State},  \country{Country}}. \email{name2@email.com}}

% \authormark{Author Name1 et al.}

% \keywords{keyword1, keyword2, keyword3, keyword4}

% \keywords[MSC Codes]{\codes[Primary]{CODE1}; \codes[Secondary]{CODE2, CODE3}}

\abstract{
This paper presents an overview of scientific modeling and discusses the complementary strengths and weaknesses of ML methods for scientific modeling in comparison to process-based models. It also provides an introduction to the current state of research in the emerging field of scientific knowledge-guided machine learning (KGML) that aims to use both scientific knowledge and data in ML frameworks to achieve better generalizability, scientific consistency, and explainability of results. We discuss different facets of KGML research in terms of the type of scientific knowledge used, the form of knowledge-ML integration explored, and the method for incorporating scientific knowledge in ML. We also discuss some of the common categories of use cases in environmental sciences where KGML methods are being developed, using illustrative examples in each category.
% Abstracts should be 250 words. It must be able to stand alone and so cannot contain citations to the paper's references, equations, etc. An abstract must consist of a single paragraph and be concise. Because of online formatting, abstracts must appear as plain as possible.
}

\end{Frontmatter}

\section{Introduction}

As advances in artificial intelligence (AI) and machine learning (ML) continue to revolutionize mainstream applications in commercial problems, there is a growing excitement in the scientific community to harness the power of ML for accelerating scientific discovery \cite{wang2023scientific,zhang2023artificial,krenn2022scientific,ai4science2023impact}. This is especially true in environmental sciences that are rapidly transitioning from being data-poor to data-rich, e.g., with the ever-increasing volumes of environmental data being collected by Earth observing satellites, in-situ sensors, and those generated by model simulations (e.g., climate model runs \cite{lynch2008origins}). Similar to how recent developments in ML has transformed how we interact with the information on the Internet, it is befitting to ask how ML advances can enable Earth system scientists to transform a fundamental goal in science, which is to build better models of physical, biological, and environmental systems. 

% [Add a paragraph summarizing Sections 2 and 3 ]
The conventional approach for modeling relationships between input drivers and response variables is to use \textit{process-based models} rooted in {scientific equations}. Despite their ability to leverage the mechanistic understanding of scientific phenomena,  process-based models suffer from several shortcomings limiting their adoption in complex real-world settings, e.g., due to imperfections in model formulations (or modeling bias), incorrect choices of parameter values in equations, and high computational costs in running high-fidelity simulations. In response to these challenges, ML methods offer a promising alternative to capture statistical relationships between inputs and outputs directly from data. However, ``black-box'' ML models, that solely rely on the supervision contained in data, show limited generalizability in scientific problems, especially when applied to out-of-distribution data. One of the reasons for this lack of generalizability is the limited scale of data in scientific disciplines in contrast to mainstream applications of AI and ML where large-scale datasets in computer vision and natural language modeling have been instrumental in the success of state-of-the-art AI/ML models. Another fundamental deficiency in black-box ML models is their tendency to produce results that are inconsistent with existing scientific theories and their inability to provide a mechanistic understanding of discovered patterns and relationships from data, limiting their usefulness in science. 

In contrast to black-box ML, a new generation of ML methods is being developed that leverage the information contained in data without ignoring the wealth of knowledge available in scientific disciplines. This emerging field of research in ML is referred to as scientific knowledge-guided machine learning (KGML) \cite{tgds} where scientific theories (e.g., the principle of mass and energy conservation) are used to guide the construction and training of ML models for improved generalizability and scientific consistency. Even though the idea of incorporating scientific knowledge in ML is relatively recent \cite{faghmous2014theory, tgds, raissi2017physics1}, this area has seen a surge in interest that is reflected in the large volume of papers published just in the last few years on topics related to KGML in a wide range of physical, biological, and environmental applications (see a recent book \cite{karpatne2022knowledge} and survey articles \cite{willard2022integrating,karniadakis2021physics,von2019informed}).

% \paragraph{Contributions and Structure of Paper:}
This paper aims to accomplish the following goals. First, it provides the ML community with a gentle introduction to the field of scientific modeling and discusses complementary strengths and weaknesses of process-based and ML models (Section \ref{sec:background}). Second, this paper provides an easy entry point to the field of KGML for a broad research audience, by summarizing prior research in the field in a simple and accessible format using a three-dimensional view, namely the type of scientific knowledge used, the form of knowledge-ML integration deployed, and the method for incorporating scientific knowledge in ML (Section \ref{sec:kgml}). Third, it provides a detailed view of three major research methods for integrating scientific knowledge in ML frameworks (Section \ref{sec:methods}). Fourth, it describes research in KGML from the context of four primary use-cases in scientific disciplines: forward modeling, inverse modeling, generative modeling, and downscaling (Section \ref{sec:use-cases}). Finally, it highlights emerging opportunities for KGML research in building foundation models in environmental science (Section \ref{sec:cross-cutting}), and provides concluding remarks and future directions of research in this field (Section \ref{sec:conclusions}). 

\begin{figure}
\centering
\includegraphics[width=\linewidth]{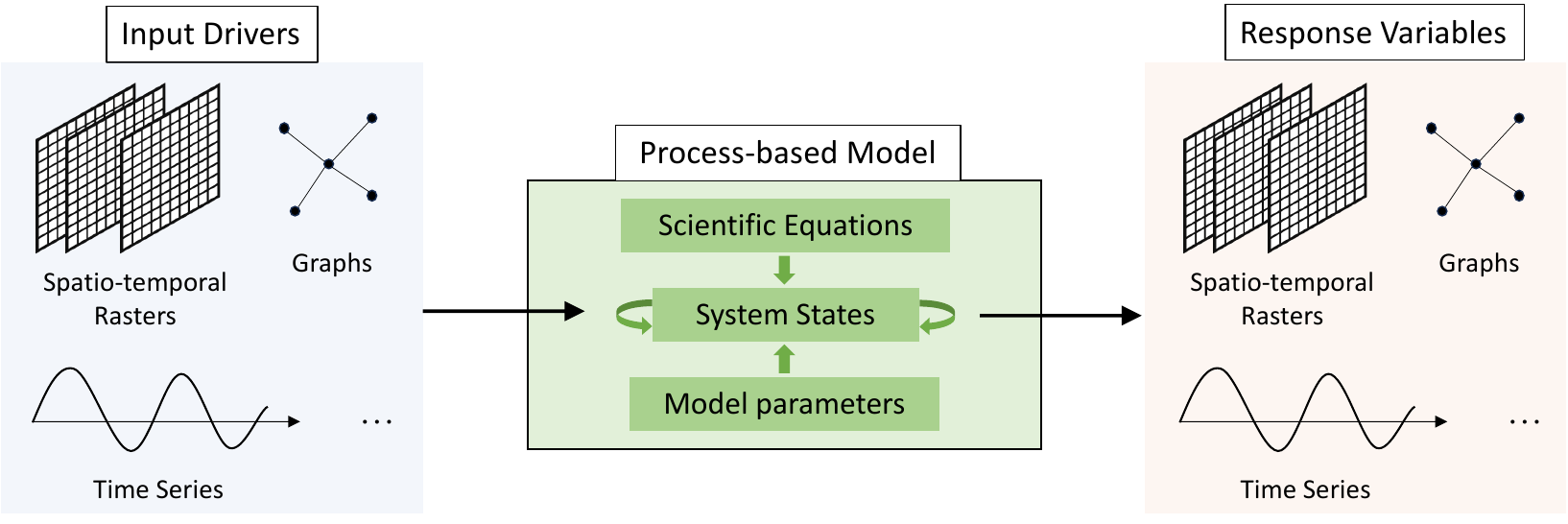}
\caption{Generic framework for scientific modeling using process-based models}
\label{fig:sc_modeling}
\end{figure}

% \subsection*{A Brief Overview of Scientific Modeling} 
\section{Background on Scientific Modeling: Opportunities and Challenges}
\label{sec:background}
One of the fundamental pursuits in science is to understand the nature of the physical world by building computational models that are capable of predicting (or simulating) the response of physical/environmental systems.
Figure \ref{fig:sc_modeling} provides a brief overview of the generic framework of scientific modeling, where the goal is to model relationships between input drivers and response variables. For example, consider the problem of modeling lake water quality. Here, we are interested in capturing relationships between input drivers such as the temperature of air above the lake and the net flow of nutrients to the lake (e.g., through agricultural discharge), and response variables of the lake such as the temperature of the water and concentrations of chemical and biological constituents such as dissolved oxygen and chlorophyll content. Note that both the input drivers and response variables can have spatial and temporal structures and can be represented as spatio-temporal rasters (available on regular grids), time-varying graphs (available on irregular grids), or time series. Environmental data for both driver and response variables are generally collected through ground-based sensors, satellites, or simulation runs of other process-based models. 
% These data sets may contain multiple modalities at different resolutions of space and time, each of which may provide partial but critical information about the underlying processes that may be highly heterogeneous, i.e., their mechanisms for interacting and evolving may vary from one region to another and over time. 
Further, while we expect the input drivers to be relatively well-observed, the response variables are generally hard to measure directly and thus have a limited number of observations at desired spatial and temporal scales.
By modeling relationships between input drivers and response variables, an important end-goal of scientific modeling is to improve our understanding of scientific systems in novel (or unseen) testing scenarios, where the distributions of input and response variables are different than those observed while building the model. 
% Environmental data collected from sensors captures information about underlying complex physical processes that conform to scientific laws (e.g., energy conservation, mass balance) and evolve and interact at different spatial and temporal scales. 

% These data sets may contain multiple modalities (e.g., optical signals vs Radar vs ..) and may be collected at different resolutions (space and time), each of which may provide partial but critical information about the underlying processes that may be highly heterogeneous in space and time (i.e., physical processes and how they interact and evolve may change from one region to another and over time).

% Why deep learning has been successful in addressing problems in a number of domains involving complex data sets with spatial and temporal (sequential) information, e.g., vision, video, and natural language processing, new advances are needed to leverage the unique nature of environmental datasets. Specifically, the ability to capture spatial and temporal context and c

% Perhaps we can create a paragraph heading: “Process-based models” and explain all about them 

% (good: mechanistic understanding, generalizaibilty; bad: necessarily approximation of reality, parameters that need to be calibrated, and inability to leverage data from highly observed regions/entities to sparsely or unobserved entities)

% by 

\subsection*{Process-based Models}
The conventional approach for scientific modeling is to use process-based models (also referred to as physics-based or science-based models), where the solution structure of relationships between inputs and response variables is rooted in {scientific equations}  (e.g., laws of energy and mass conservation). In particular, process-based models make use of scientific equations to infer the evolution of time-varying latent or hidden variables of the system, also referred to as system states (see Figure \ref{fig:sc_modeling}). Additionally, process-based models often involve {model parameters} that need to be specified or calibrated (often with the help of observational data) for approximating real-world phenomena. For example, state-of-the-art models in  {lake modeling} such as the general lake model (GLM) \cite{hipsey2014glm} involve parameters such as water clarity that can dramatically change the nature of relationships between inputs and outputs and have to be tuned for every lake based on observations of input-output pairs. 
% Should this be moved to the place where we discuss the limitations of process models?  We can even expand it further.   Perhaps here we can expand on the mechanistic understanding component, which is easy for people to see. 
% Furthermore, while we have may access to some input drivers influencing target variables, there may be many other auxiliary variables affecting the system states that are not fully observed at the required resolutions of space and time. For example, in hydrology, we may be able to infer the type of vegetation or the amount of moisture in the soil with the help of data collected by Earth-orbiting satellites, which can provide valuable information about the state of hydrological processes operating at a given region of the Earth at a certain time. However, the mapping of satellite observations to the hidden system states of hydrology may not be fully known, or we may not have full coverage of satellite observations at the required spatial and temporal resolutions. Techniques developed in the area of data assimilation attempt to resolve how real-time data is used to update the evolution of system states in process-based models. 
Taken together, the conventional paradigm of scientific modeling---developed over centuries of systematic research---forms the foundation of our present-day understanding of scientific systems across a wide spectrum of problems in environmental sciences.

% \subsection*{Goals of Scientific Modeling}
A key feature of process-based models is their ability to provide a mechanistic understanding of the cause-effect mechanisms between input and output variables that can be used as a building block for advancing scientific knowledge. As a result, process-based models are continually updated and improved by the scientific community to fill knowledge gaps in current modeling standards and discover new theories and formulations of scientific equations that better match with observations and are scientifically meaningful and explainable. Since process-based models are rooted in scientific equations that are assumed to hold true in any testing scenario, they are also expected to easily generalize even outside the data used during model building.  For example, process-based models can be made to extrapolate in space (e.g., over different geographic regions), in time (e.g., forecasting the future of a system under varying forcings of input drivers), or in scale (e.g., discovering emergent properties of larger-scale systems using models of smaller-scale system components).

% to generalize our understanding of the behavior of the scientific system on novel (or unseen) testing scenarios, We may also want to extrapolate in scale (e.g., by building a model that works on 10 to 20 lakes, can we understand the dynamics of lakes at continental scales?).

\subsection*{Limitations of Process-based Models}

Despite the widespread use of process-based models in almost every domain of science, they suffer from certain research gaps limiting their effectiveness in many real-world scientific applications.  \textit{First}, process-based models often involve several parameters that have to be calibrated for every new system using observations specific to the system. For example, in the problem of modeling the quality of water in lakes, there are several hundreds of parameters in process-based models such as GLM-Aquatic EcoDynamics (GLM-AED) \cite{hipsey2019general} capturing different aspects of lake thermodynamics and chemical/biological processes that need to be calibrated for every lake using observational data. While it is still possible to calibrate model parameters in highly-observed systems with a sufficient number of data samples, process-based models are difficult to generalize in sparsely-observed systems where the underlying model parameters can be difficult to estimate using limited (or sometimes non-existent) observations. For example, we have a sufficiently good number of observations in a few highly-monitored lakes but for the vast population of lakes in the U.S. and across the world, we do not have dedicated programs for high-resolution monitoring of lake water quality, making it challenging to calibrate process-based models with little to no data. 

\textit{Second}, process-based models frequently {suffer from several imperfections} (or bias) in their solution structure (e.g., due to incomplete understanding of the system or approximations introduced in the model to speed up model runs), leading to poor generalizability of model outputs. For example, while state-of-the-art models in lake modeling such as the GLM \cite{hipsey2014glm} adequately capture the thermodynamic processes responsible for mass and energy balance in lakes, the processes governing the biogeochemistry of lake nutrients such as dissolved oxygen, organic matter, and phytoplankton counts are less well-understood. Such processes are generally approximated using over-parameterized forms of weakly-informed equations, introducing large modeling errors and uncertainties. Furthermore, the introduced over-parametrizations are often difficult to understand and relate to physically meaningful quantities, making them fall short in serving the goal of explainability.

\textit{Third}, process-based models are often {computationally expensive} to be run at the required scales in operational settings. In particular, in many scientific problems, learning the `forward' mapping from input drivers to response variables (also referred to as the problem of {forward modeling}) requires finding solutions to complicated scientific equations often involving complex numerical operations. On the other hand, learning the `inverse mapping' from observed inputs and outputs to model parameters (referred to as the problem of {inverse modeling}) is also a computationally demanding process as it requires multiple runs of the forward model for every round of model inversion \cite{tarantola2005inverse,ljung1998system,beven1992future}. For example, in the problem of lake modeling, custom-calibrating the GLM model is both labor- and computation-intensive, and hence there is a trade-off between increasing the accuracy of GLM estimates and expanding the feasibility of study to a large number of lakes. 

% A third goal is to go from fully observed to less-observed systems. [Discuss how this is challenging since the model parameters have to be calibrated for every new system in the conventional framework. However, we may not have sufficient data in the target system to calibrate the parameters. Is it possible to share the learning from related systems (e.g., by borrowing default settings of model parameters?)]

% Finally, a fourth goal is to identify gaps in current state of process-based models in regimes where we know they are lacking and devise ways to augment or replace them. For example, if we know a set of equations are not complete, we can introduce additional parameterizations. However, the new parameters should be explainable so that a domain scientist can understand their physical relevance.

% 1. Generalize. 2. Need parameters. 3. May not be complete. 4. Efficiency (sample, compute, parameter). 5. Explainability. 

% OLD LIST (TO BE REMOVED):

% \begin{enumerate}
%     \item Out of sample generalization (interpolation vs extrapolation). 
%     \item How do we go from highly observed to less observed systems?
%     \item Can we go beyond the goal of improving generalization performance to improving explainability of discovered patterns/models/relationships in a manner consistent with existing scientific theories?
%     \item Incorporating Physical laws
%     % \item Improving the parameter efficiency and training/testing speed of ML models, along with generalizability
% \end{enumerate}

\subsection*{Opportunities for ML in Scientific Modeling}
To address the research gaps of process-based models in scientific modeling, ML methods offer a promising alternative to capture relationships between input drivers and output response variables directly from data. 
% For example, ML methods can act as fast {surrogates} of expensive process-based models or improve the accuracy of imperfect models by learning generalizable patterns from paired examples of input and output variables. 
ML models are similar to process-based models in many ways. For example, both ML and process-based models pursue the common goal of mapping inputs to outputs. Just like the parameters of process-based models, many ML models also involve several learnable parameters (e.g., the weights and biases of deep neural networks) that have to be inferred from data often using computationally efficient optimization techniques such as first-order gradient descent. 

However, ML models also show fundamental differences from process-based models endowing them with complementary advantages. \textit{First}, while the functional forms of process-based models may have imperfections that are difficult to fix using data, the solution structure of ML models (e.g., deep learning models) use generic architectures that are highly expressive and capable of capturing arbitrarily complex relationships between inputs and outputs provided they are fed with adequate training data. \textit{Second}, ML methods (e.g., deep learning methods) are very fast at making forward inferences using simple matrix operations that are amenable to modern GPU architectures. As a result, ML models, once trained, can offer significant computational gains in modeling complicated scientific systems in comparison to process-based models employing complex numerical operations. \textit{Third}, a unique advantage of many ML frameworks (e.g., deep learning) is that the parameters of ML models trained over a certain system can be used as a starting point to initialize the learning of parameters in a new system. A number of research strategies have been developed over the years in ML in the fields of transfer learning and meta learning to enable the transfer of learning from source tasks with a sufficient number of data samples to target tasks with little to no data, possibly even with shifts in distributions of input and output variables. Building upon the idea of transfer learning in mainstream ML applications of computer vision and natural language understanding, there is a growing trend to create Foundation models \cite{bommasani2021opportunities} such as chatGPT \cite{openai2023gpt} that have been pre-trained on a wide variety of tasks to learn universal feature representations of data, which can be fine-tuned on a new task with very few data. This inherent transferability of learning in ML models provides a novel opportunity in environmental sciences to build broad-scale models that leverage knowledge from a large number of well-observed systems and can be fine-tuned on new less-observed systems with very few (or even no) observations. 
% \textcolor{red}{Fourth, the intricate structure of the ML model allows the information to be forwarded through  numerous branches of routes in the model, facilitating the encoding of both shared and distinct information essential for various tasks. %there are many branches of routes within the the complex ML model structure, which enables encoding  shared and distinct information needed for multiple tasks. 
% The power of ML models in addressing multiple tasks is also validated by the recent success of foundation models in a variety of tasks without the need to change the model settings. This provides new opportunities for developing unified data-driven models that can handle scientific modeling tasks over different locations and  different time periods, and  perform predictions at different scales and for different target variables specified by the user. }
% [Mention some systems are highly observed, some are less observed. Transfer learning has been going on well before foundation models.]

% potentially transferred to other systems. [Mention the advances in the field of transfer learning and Foundation models, where we are able to learn universal features that can be fine-tuned on new tasks even with little data. This 

\subsection*{Challenges for Black-box ML}
Despite their promise, black-box ML models---that are designed, trained, and deployed agnostic to scientific theories---have led to spectacular failures in scientific problems \cite{caldwell2014statistical,Lazer2014,marcus2014eight}(e.g., the rise and fall of Meta AI's Galactica \cite{galactica} and Google Flu Trends \cite{Lazer2014}). 
This is primarily because black-box ML models solely rely on the supervision contained in data and hence are only as good as the data they are trained with. In problems where we have limited coverage of data distributions in the training set, black-box ML models can fail to generalize on out-of-distribution samples and produce results that are not only inaccurate but also nonsensical and incoherent with existing scientific theories while being highly overconfident. This is in contrast to process-based models that are designed to produce robust and scientifically consistent predictions even if they are inaccurate due to poor calibration of parameters or modeling imperfections. One strategy to improve the generalization capability of black-box ML models is to scale the size and diversity of training data along with increasing the size of models, which is the approach pursued in the creation of Foundation models \cite{bommasani2021opportunities}. While this is possible in mainstream applications of ML where we have Internet-scale data with millions \cite{imagenet_cvpr09} or even trillions \cite{norvig2009natural} of labeled samples, the nature of scientific data is very different that makes it infeasible to match the scales of labeled data possible in commercial domains. In particular, scientific systems exhibit rich heterogeneity in the distributions of input and output variables over space and time, with complex interactions of processes operating at multiple scales with varying memory footprints, often involving unknown confounding factors. This makes it difficult to cover the full spectrum of possible data distributions that one may encounter when deploying ML models in scientific applications. 

Another fundamental limitation of black-box ML models is their inability to generate \textit{explainable} hypotheses of the relationships between input and output variables in a manner consistent with existing scientific theories. Since one of the primary goals in science is to explain the cause-effect mechanisms of phenomena observed in the physical world, a black-box model that achieves somewhat good predictive performance but fails to deliver a mechanistic understanding of the underlying processes is unfit for being used as a basis for subsequent scientific developments. {There is a growing interest in machine learning to incorporate the tools of causal inference \cite{pearl2018book,pearl2019seven} to move beyond learning associations in data (which is what most predictive models do) to designing interventions and eventually reasoning with counterfactuals. Scientific problems provide a unique opportunity to study causal reasoning rooted in existing theories of cause-effect mechanisms available as process-based models.}
% (Note that this is in stark contrast to black-box ML models that solely rely on the information contained in the training data and hence fail to generalize on out-of-distribution data samples.) 

\section{Knowledge-guided Machine Learning (KGML): An Overview}
\label{sec:kgml}
\begin{figure}
\centering
\includegraphics[width=\linewidth]{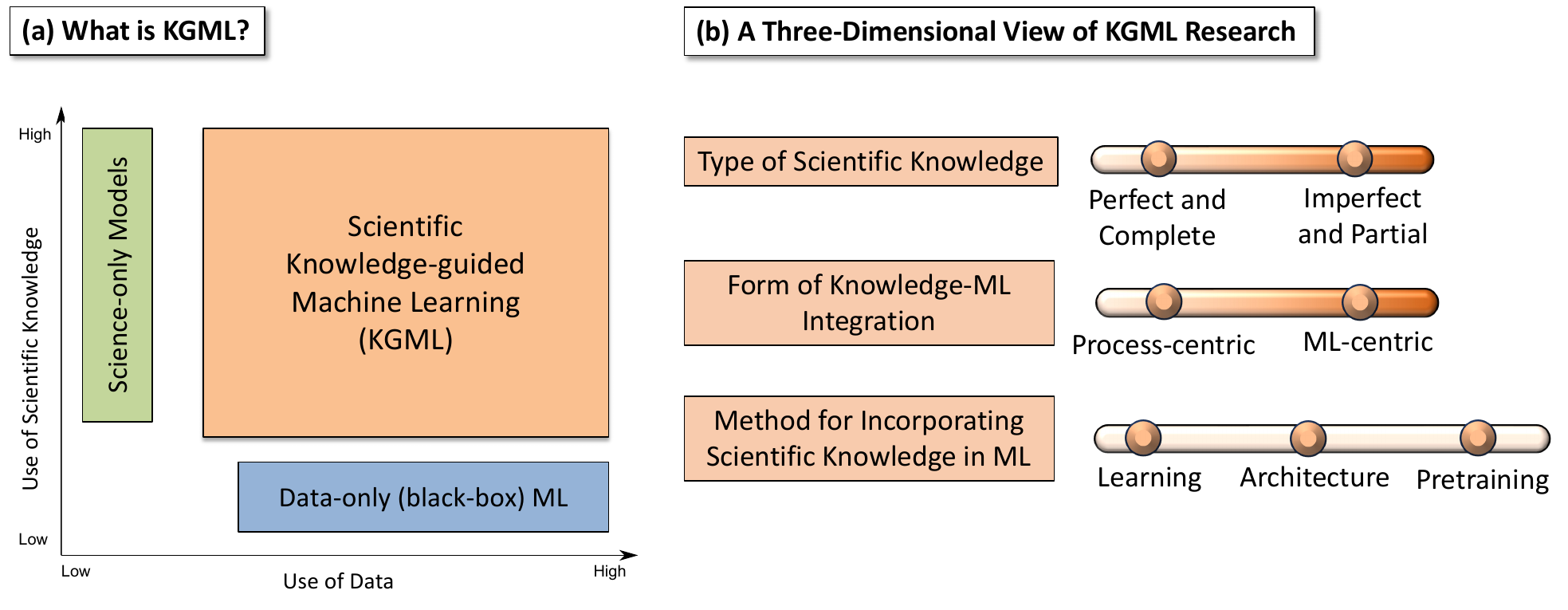}
\caption{(a) Pictorial depiction of how KGML models are different from data-only (black-box) ML models and science-only models. (b) A three-dimensional view of KGML research where different settings of these dimensions correspond to various research directions in KGML}
\label{fig:kgml_overview}
\end{figure}

Despite the challenges in using black-box ML, modeling of environmental systems also presents novel opportunities for developing new advances in ML that leverage the unique characteristics of data encountered in scientific domains. For example, data collected in environmental sciences capture valuable information about a variety of Earth system processes operating at different spatial and temporal scales. While a given data set may only provide a partial view of the complex nexus of physical processes evolving over space and time, they represent real-world phenomena grounded in core scientific theories (e.g., the principle of mass and energy conservation). As a result, there is a growing interest in scientific and ML communities to leverage {scientific knowledge} in ML frameworks to produce explainable and scientifically grounded solutions that are likely to generalize even on out-of-distribution samples with limited training data. Research in this emerging field is referred to as {scientific knowledge-guided machine learning} (KGML) \cite{karpatne2022knowledge,tgds} where both scientific knowledge and data are used at an equal footing as complementary sources of supervision in ML frameworks, in contrast to first-principle process-based models and data-only (black-box) models (see Figure \ref{fig:kgml_overview}(a)). 
% \subsection*{Overview of Previous Research in KGML}
KGML is a rapidly growing research community that has seen several notable examples of promising research in recent years.  One of the first major efforts to present the roadmap for research in KGML includes a 2017 perspective article \cite{tgds} that laid the foundation of several research methods for combining scientific knowledge in ML framework. This consequently sparked ample interest in the ML and scientific communities to pursue research in KGML, as evidenced by the large volume of literature published in the last few years including survey articles \cite{willard2022integrating,karniadakis2021physics,von2019informed}, perspective papers \cite{reichstein2019deep,xu2023physics}, and a recent book on KGML \cite{karpatne2022knowledge}. 
% A recent survey article  by Willard et al. \cite{willard2022integrating} in 2021 covered 350+ papers published on the topic of KGML, many of which were published after 2017. 

% Each  provide partial but critical information about the underlying processes that may be highly heterogeneous in space and time (i.e., physical processes and how they interact and evolve may change from one region to another and over time).
% Just as advances in deep learning have been successful in addressing problems in a number of mainstream applications involving data sets with spatial and temporal (sequential) information, e.g., vision, video, and natural language processing, new advances are needed to leverage the unique nature of environmental datasets. Specifically, the ability to capture spatial and temporal context and c

% \begin{enumerate}
%     \item Generalizability with limited labeled data. Non-sensical behavior.
%     \item Heterogeneity in space and time
%     \item Complex Processes (multi-scales, not all variables observed, show the figure of inputs to outputs with hidden knowledge of underlying processes)
%     \item Lack of scientific consistency
%     \item Black box models unfit for discovering scientific knowledge
    
% \end{enumerate}

\subsection*{A Multi-dimensional View of KGML Research}
KGML has quickly grown to become a thriving area of research that is reflected in the diversity of research communities that have emerged to leverage KGML  in the context of varied scientific problems. As a result, research in KGML is also referred to by many names including  `theory-guided data science', `physics-guided machine learning', `science-guided machine learning', and `physics-informed machine learning'. While all of these terms refer to the common goal of integrating scientific knowledge with ML, it is important to understand the differentiating aspects of KGML research that lead to its rich diversity. We present these differentiating aspects in the form of a three-dimensional view of prior research in KGML: type of scientific knowledge (ranging from perfect and complete to imperfect and partial), form of knowledge-ML integration (ranging from process-centric to ML-centric), and method for incorporating scientific knowledge in ML (with choices including knowledge-guided learning, architecture, or pretraining). See Figure \ref{fig:kgml_overview}(b) for a schematic depiction of these three dimensions, which represent different steps or decisions one must take to take to select a KGML formulation best-suited for a scientific problem. They also help in categorizing previous research in KGML and exposing commonalities between research directions explored in diverse scientific communities. 
We briefly describe each of these three dimensions in the following.
% In the following, we highlight some prominent examples of prior research in KGML.

 \subsubsection*{Type of Scientific Knowledge} The first aspect that one must decide when working on a KGML problem is the type of scientific knowledge available in the problem. Scientific knowledge is available in diverse formats in various disciplines, including first-principle equations, domain-specific laws, rules, heuristics, or ontological relationships. In many scientific applications, we only have partial knowledge of the system (i.e., we may not have complete knowledge of all system processes) or we have to deal with approximate terms in scientific equations. We refer to such systems as having \textit{partial} and/or \textit{imperfect} scientific knowledge.
 A seminal line of work in incorporating partial and imperfect scientific knowledge in the learning of ML models is the framework of {physics-guided neural networks (PGNNs)} \cite{karpatne2017physics} that was originally developed in 2017 for modeling the temperature of water in lakes using monotonic constraints of density and depth. This has led to a growing body of research in KGML for incorporating diverse forms of scientific knowledge including monotonic constraints \cite{muralidhar2018incorporating}, eigenvalue equations \cite{elhamod2022cophy,ghosh2022physics}, energy balance \cite{jia2019physics,jia2021physics,read2019process}, and scaling laws \cite{hanson2020predicting}. For example, Jia et al. developed the framework of physics-guided recurrent neural networks (PGRNNs) \cite{jia2019physics} that accounted for the conservation of energy across time for predicting lake temperatures, while allowing for error tolerance in the equation for energy balance involving approximate terms. 
 
 In other applications,  scientific knowledge can be assumed to be \textit{perfect} and \textit{complete}, i.e., all of the underlying processes of a scientific system can be fully described by a set of scientific equations that are exactly known with no incorrect or missing terms. For example, in many applications, scientific knowledge is expressed in the form of idealized partial differential equations (PDEs), where the goal is to solve these equations in a computationally efficient manner. Many techniques in KGML have been developed for problems where scientific knowledge can be assumed to be perfect and complete. 
% In some problems, we can assume that the underlying processes of a scientific system are fully described by a given set of scientific equations (e.g., PDEs),  and the goal is to solve these equations in a computationally efficient way to model scientific systems. 
% For example, the flow of fluids in idealized systems is completely determined by the Navier--Stokes equations. In such problems, even though we have perfect knowledge of the equations, solving them numerically at required scales may not be computationally feasible. 
In such problems, ML methods can serve as a promising alternative to numerical methods for solving scientific equations with significant computational gains. One of the seminal works in this direction is the framework of  Physics-Informed Neural Networks (PINNs) \cite{raissi2017physics1,raissi2017physics2,raissi2019physics} that were developed in 2017 to solve PDEs using feed-forward neural networks. This has opened an entirely new line of research on solving PDEs using advanced neural architectures such as neural operators \cite{wang2021learning,li2021physics,goswami2022physics}. 
There is thus a continuum of research directions in KGML based on the type of scientific knowledge, which can range from perfect and complete to imperfect and partial.

% REPHRASE:[ They use loss functions to minimize PDE residual loss for a fixed initial/boundary condition. This started a new line of research on solving PDEs with neural networks. A recent addition is a new class of neural operators that can solve a family of PDEs over a range of initial/boundary conditions. Limitations: works with idealized physics, assumes physics knowledge is perfect and complete].

% [However, many real-world problems in science involve scientific knowledge that is neither perfect nor complete, where PINNs cannot be applied direclty. Here, physics-guided loss functions only capture a partial view of the reality. The loss functions may also have approximations. There is an opportunity to use physics loss along with data loss. There is also a value of using simulation data for pretraining models, along with using observational data for fine-tuning. This is the framework of PGNNs and PGRNNs for ecology. Also mention hybrid modeling.]

\subsubsection*{Form of Knowledge-ML Integration} There is also a continuum of research in KGML depending on the exact form of knowledge-ML integration, ranging from ML-centric approaches to process-centric approaches. In ML-centric approaches, the mapping of inputs to response variables is primarily driven by ML methods while scientific knowledge plays the role of guiding ML algorithms to scientifically consistent solutions. We describe the different strategies for incorporating scientific knowledge in ML frameworks in detail in Section \ref{sec:methods}. On the other hand, in process-centric approaches, the entire modeling exercise is primarily driven by process knowledge whereas ML methods are only used to inform or augment certain components of process-based models. In these approaches, ML is embedded inside the process-based model and the final predictions of response variables are made by the process-based model. For example, the role of ML can be limited to identifying (or \textit{calibrating}) latent parameters in process-based models from data (e.g., in a recent line of work on differential parameter learning for hydrology \cite{shen2023differentiable}), which results in better match between process-based model outputs with observations of response variables. Another example of process-centric KGML formulations is the area of subgrid parameterization in climate science \cite{bolton2019applications}, where ML methods are used to learn corrective terms in science-based equations running at coarse resolutions of space (e.g., at 1km grids), to capture the effect of subgrid processes at finer resolutions not accounted by the equations. This is related to the general direction of using ML to augment certain components in process-based models that are known to be imperfect, e.g., in the field of turbulence modeling to close the gap between high-fidelity and low-fidelity simulations \cite{duraisamy2019turbulence} and in geosciences where ML-augmented equations are used to model hydrological processes \cite{feng2022differentiable}. There is also a range of approaches in KGML to create hybrid combinations of ML-based and process-based components, referred to as the field of hybrid modeling \cite{tgds, karpatne2022knowledge,willard2022integrating}. Approaches in this field include residual modeling strategies \cite{thompson1994modeling}, where ML is simply used to learn the residuals of process-based models and the final predictions are equally contributed by both ML and process-based model outputs. Another strategy is to use the outputs of process-based models as additional inputs in ML models along with drivers to predict response variables, referred to as the framework of hybrid-physics-data (HPD) models \cite{karpatne2017physics}. This style of hybrid modeling is closer to being ML-centric as the final predictions of response variables are determined by the ML model, and the process-based simulations are only used as inputs that serve as initial estimates. 
% Finally, there are several strategies for incorporating scientific knowledge in ML frameworks, as described in the following.

% or learning correction terms in models

% (also known as the problem of model calibration \cite{})

% provided by a 

% [Knowledge-centric: ML embedded in scientific models: parameter calibration. dPL. subgrid parameterization. ML-centric: knowledge embedded in ML models. Hybrid modeling has a continuum.]

\begin{figure}
\centering
\includegraphics[width=0.85\linewidth]{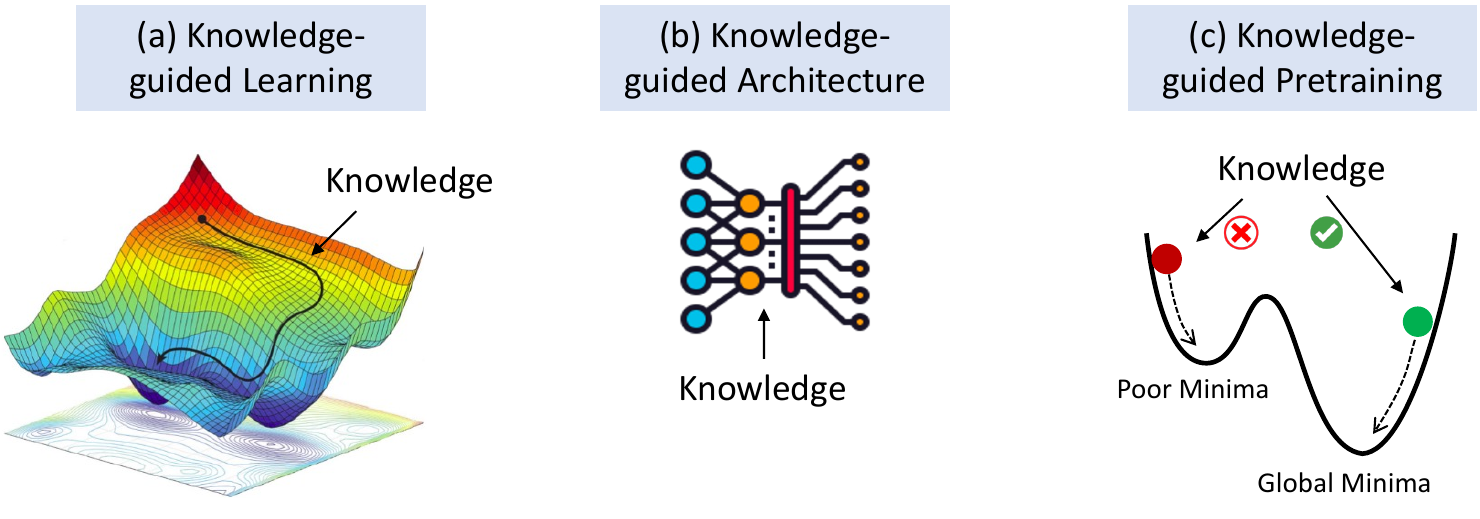}
\caption{Research methods for incorporating scientific knowledge in ML frameworks}
\label{fig:kgml_methods}
\end{figure}

\subsubsection*{Method for Incorporating Scientific Knowledge in ML} 
When it comes to using ML components in KGML approaches (either process-centric or ML-centric), an important aspect is the method used for incorporating scientific knowledge in ML frameworks. One of the 
primary ways of doing this is by modifying the \textit{learning} algorithm of ML models (e.g., deep learning models) to favor the selection of models that are consistent with scientific equations (e.g., using knowledge-guided loss functions), so as to steer the training trajectory toward generalizable solutions. This method is adopted in the frameworks of PGNN and PGRNN for lake temperature modeling and the framework of PINNs for solving PDEs. Another approach is to   ``hard-code'' or ``bake-in'' scientific knowledge directly in the solution structure of deep learning models, resulting in knowledge-guided neural network architectures. Some examples of research in this area include the framework of physics-guided architecture (PGA) of LSTM models (PGA-LSTM) \cite{daw2020physics} and approaches for capturing symmetries and invariances of dynamical systems in the architecture of neural networks \cite{wang2020incorporating}. The basic goal of these methods is to improve the scientific explainability of features learned at the hidden layers of neural networks, grounded in scientific theories. A third approach is to initialize neural network parameters using weights informed by scientific knowledge, e.g., by pretraining the model on simulations of science-based models before fine-tuning on gold-standard observations \cite{jia2019physics,jia2021physics}. See Figure \ref{fig:kgml_methods} for a pictorial depiction of these three methods. Note that these three methods are also not mutually exclusive; they can be combined together in various ways depending on the context of application.

\begin{table}[tbh]
\centering
\begin{tabular}{|c|c|c|c|c|}
\hline
\textbf{\begin{tabular}[c]{@{}c@{}}KGML \\ Research\end{tabular}} &
  \textbf{\begin{tabular}[c]{@{}c@{}}Type of \\ Knowledge\end{tabular}} &
  \textbf{\begin{tabular}[c]{@{}c@{}}Knowledge-ML \\ Integration\end{tabular}} &
  \textbf{\begin{tabular}[c]{@{}c@{}}KGML \\ Method\end{tabular}} &
  \textbf{\begin{tabular}[c]{@{}c@{}}Additional\\ Comments\end{tabular}} \\ \hline
PINN \cite{raissi2019physics} &
  \begin{tabular}[c]{@{}c@{}}Perfect \& \\ Complete\end{tabular} &
  ML-centric &
  Learning &
  \begin{tabular}[c]{@{}c@{}c@{}}Solves a single PDE\\ at any arbitrary point \\ using feedforward networks \end{tabular} \\ \hline
\begin{tabular}[c]{@{}c@{}}Neural Operators \\ (FNO \cite{li2020fourier}, \\ DeepONet \cite{lu2021learning})\end{tabular} &
  \begin{tabular}[c]{@{}c@{}}Perfect \&\\ Complete\end{tabular} &
  ML-centric &
  \begin{tabular}[c]{@{}c@{}}Learning \\ (optional)\end{tabular} &
   \begin{tabular}[c]{@{}c@{}c@{}}Solves a family of PDEs\\ at arbitrary resolutions \\ using feedforward networks \end{tabular} \\ \hline
PGNN \cite{daw2022physics} &
  \begin{tabular}[c]{@{}c@{}}Imperfect \&\\ Partial\end{tabular} &
  ML-centric &
  Learning &
  \begin{tabular}[c]{@{}c@{}c@{}c@{}}Uses monotonic constraints \\ \textit{(partial)} and  process \\ simulations (\textit{imperfect}) \\ in feedforward networks \end{tabular} \\ \hline
PGRNN \cite{jia2019recurrent} &
  \begin{tabular}[c]{@{}c@{}}Imperfect \&\\ Partial\end{tabular} &
  ML-centric &
  \begin{tabular}[c]{@{}c@{}}Learning \&\\ Pretraining\end{tabular} &
  \begin{tabular}[c]{@{}c@{}c@{}c@{}}Uses energy conservation \\ (\textit{imperfect and partial})  \\ in RNN with process \\  simulations (\textit{imperfect})\end{tabular} \\ \hline
PGA-LSTM \cite{daw2020physics} &
  \begin{tabular}[c]{@{}c@{}}Perfect \&\\ Partial\end{tabular} &
  ML-centric &
  Architecture &
  \begin{tabular}[c]{@{}c@{}c@{}}Uses monotonic constraints\\ (\textit{perfect and partial}) \\ in LSTM architecture\end{tabular} \\ \hline \begin{tabular}[c]{@{}c@{}} Equivariant-Nets \\\cite{wang2020incorporating}\end{tabular}
  &
  \begin{tabular}[c]{@{}c@{}}Perfect \&\\ Partial\end{tabular} &
  ML-centric &
  Architecture &
  \begin{tabular}[c]{@{}c@{}}Uses symmetries (\textit{perfect} \\ and \textit{partial}) in CNN\end{tabular} \\ \hline
dPL \cite{shen2023differentiable} &
  \begin{tabular}[c]{@{}c@{}}Perfect \&\\ Complete\end{tabular} &
  Process-centric &
  - &
  \begin{tabular}[c]{@{}c@{}}Learns parameters in \\ differentiable process model\end{tabular} \\ \hline
\begin{tabular}[c]{@{}c@{}c@{}}Subgrid \\ Parameterization \\ \cite{bolton2019applications} \end{tabular} &
  \begin{tabular}[c]{@{}c@{}}Imperfect \&\\ Complete\end{tabular} &
  Process-centric &
  - &
  \begin{tabular}[c]{@{}c@{}}Uses ML to correct\\ effects of subgrid processes\end{tabular} \\ \hline
\end{tabular}
\caption{Categorization of previous research in KGML in terms of the three-dimensional view of KGML research}
\label{tab:KGML_categorization}
\end{table}

% \textcolor{red}
{Table \ref{tab:KGML_categorization} provides a categorization of a few example research works in KGML from the perspective of the three dimensions. They are described in more detail in later sections.}

\section{Research Methods in KGML}
\label{sec:methods}

In this section, we expand upon the three primary ways of integrating scientific knowledge in ML that we introduced in the previous section, namely knowledge-guided learning algorithms (e.g., using loss functions), knowledge-guided architecture of ML models, and knowledge-guided pretraining or initialization of ML models.

\subsection{Knowledge-guided Learning}
% % \begin{enumerate}
% %     \item How can we learn with limited labeled data in scientific problems? How can we use scientific knowledge as another form of supervision for designing/training ML models? Can we leverage self-supervised/semi-supervised/distantly-supervised/weakly-supervised paradigms? Discuss PINNs, PGNNs, PGRNNs, ...

% %     \item How can we construct knowledge-guided loss functions for different forms of scientific knowledge (e.g., equations, heuristics, rules, ontologies, knowledge graphs)? Lots of work in solving PDEs. How do we handle imperfect knowledge? What if knowledge is not differentiable? Can we leverage developments in neuro-symbolic AI for differentiable reasoning in science?  

% %     \item What optimization challenges do we anticipate when using knowledge-guided loss functions (e.g., competing objectives, presence of local minima), and how can we identify/visualize/fix them? Issues due to imperfect data for constraints/laws/equations.

% %     \item Factuality checking in LLMs. Is this related to checking scientific consistency?
% % \end{enumerate}

% \textcolor{red}{Shengyu}

%In environmental science problems, where data is often limited, leveraging domain-specific knowledge from physics in machine learning models can enhance accuracy and robustness. 
Directly building ML models from limited observational data often leads to the learning of spurious patterns that cannot be generalized to new scenarios, which necessitates the awareness of underlying physical knowledge in the learning of ML models. 
The predominant method for integrating scientific knowledge into ML is to directly introduce physical principles into the training objective of ML models~\cite{karpatne2022knowledge,karpatne2017physics,chen2024hossnet,jia2019physics,raissi2019physics}.  %into the ML models for building physics-guided neural networks. 
%Various methods have emerged under different learning settings. In semi-supervised learning~\cite{yang2022survey}, physical constraints can guide the use of unlabeled data, ensuring model outputs on unlabeled data remain consistent with established physical laws~\cite{bhullar2023simultaneous, taghizadeh2022semi}. %Self-supervised learning~\cite{liu2021self} can be designed around tasks that inherently follow physical principles~\cite{hoffmann2023atmodist}, such as predicting future states based on known conservation laws. 
%Weakly-supervised learning~\cite{zhou2018brief} can use bounds or limits derived from physics as a form of guidance for models~\cite{li2020accurate, vongkusolkit2022physics}. Lastly, distantly-supervised learning also offers a possible direction for integrating auxiliary datasets with physics laws to provide additional training information~\cite{cai2022coarse}. 
%By seamlessly integrating physical knowledge, machine learning models can be better equipped to handle the challenges in environmental science problems. 
For example, the framework of physics-guided neural network (PGNN) \cite{karpatne2017physics}  was one of the first works to add physical constraints (in particular, the density-depth relationship) as loss functions to guide the training of neural networks to physically consistent solutions in the context of predicting lake water temperature. 
% bridge data-driven insights with physical constraints, making them versatile for diverse environmental science tasks, from weather forecasting to atmospheric studies. 
The PGRNN model~\cite{jia2019physics} further extended this idea to capture the heat transfer process as physics-guided loss functions in the prediction of water temperature dynamics in lakes. This method represents a pioneering effort in integrating the energy conservation law into ML training. In particular, this method estimates intermediate heat fluxes in the heat transfer process and then defines the energy conservation over these fluxes and the predicted water temperature over the water column. 
Read et al. ~\cite{read2019process}, for the first time, systematically evaluated the generalizability of the PGRNN approach on out-of-distribution samples. They demonstrated that the PGRNN model significantly outperforms both standard ML models and process-based models when tested on years or seasons with very different weather conditions compared to training data.  
%integrates the physics-based constraint over both predicted variables and intermediate fluxes into the loss function to avoid predictions that violate the energy conservation laws.  %, making it invaluable for scenarios like fluid flow simulations. 
%Meanwhile, PGNNs~\cite{daw2017physics} bridge data-driven insights with physical constraints, making them versatile for diverse environmental science tasks, from weather forecasting to atmospheric studies. 

% Scientific knowledge comes in different forms, and the 
The idea of knowledge-guided learning can be utilized to integrate scientific knowledge in different forms as loss functions in the training of ML models. % can be considered in enhancing the training loss function. %  knowledge-guided loss functions. The objective is to integrate domain-specific knowledge with data-driven methods 
%with the aim to enhance model accuracy and reliability~\cite{karpatne2022knowledge, liu2022kgml, pei2023applying}. %The scientific knowledge comes in different forms. For instance, 
For example, scientific theories are commonly represented as equations, particularly differential equations, which offer a direct way to embed known scientific laws into models~\cite{luo2023physics} such as the approach of physics-informed neural networks (PINN) \cite{raissi2019physics} and the physics-informed neural operator (PINO)~\cite{li2021physics}. %The PINN approach~\cite{raissi2019physics}, on the other hand, is introduced to solve PDEs using a feed-forward neural network while regularizing it by adding the PDEs into the training loss function.
Integrating scientific equations into the loss functions of ML models can be especially useful when the goal is to solve a known target equation, e.g., in the problem of solving PDEs%, are able to guide the model, especially in scenarios like solving PDEs
~\cite{arif2022computational, yang2023fourier, boussif2022magnet}. Similarly, heuristics, rules, and structured relationships in ontologies and knowledge graphs can also be encoded and added to the loss function~\cite{wu2022linkclimate,wu2023improving, wu2021ontology, fisher2023opening,chen2024reconstructing}. By penalizing deviations from these established knowledge bases, the training algorithm has a higher chance of learning physically consistent data patterns and thus achieving improved generalizability. % predictions become more aligned with domain-specific truths. 

However, there are challenges in defining and utilizing knowledge-guided loss functions when dealing with imperfect knowledge, e.g., equations not perfectly aligned with true observations. In such cases, we can adjust the weight of regularization terms in the loss functions, enabling a balance between minimizing training errors and satisfying domain constraints~\cite{karniadakis2021physics, raissi2019physics, li2021physics}. 
Moreover, incorporating knowledge-guided loss functions %introduces many challenges in optimization. A significant challenge is the 
can introduce potential conflict between data-driven and domain-guided loss functions, which can lead to the challenge of competing objectives during ML training \cite{elhamod2022cophy}. 
% The model might struggle with balancing the need to follow data patterns and adhere to the imposed domain knowledge. 
Additionally, the optimization landscape could suffer from the presence of poor local minima due to the added complexity of knowledge constraints, making convergence to an optimal solution more elusive \cite{wang2021understanding,daw2023mitigating,wang2022improved,wang2023expert}. 
%Multiple strategies could be possibly used to address the 
%Such conflicts amongst different objectives in optimization 
%Several strategies could be possibly used to address these challenges in optimization
These issues can be potentially addressed by multi-objective optimization strategies~\cite{guo2022knowledge, karpatne2022knowledge, kumar2023advanced}, %. Multi-objective optimization~\cite{zuo2023process} 
which offer a solution by exploiting trade-offs between conflicting goals. %~\cite{guo2022knowledge, karpatne2022knowledge}. %Additionally, due to the complex pattern of data, which can be sparse or noisy, regularization techniques become essential to prevent overfitting and ensure models generalize well to real-world scenarios~\cite{kumar2023advanced}.

\subsection{Knowledge-guided Architecture}
% \textcolor{red}{Three stages: (1) embed physics into specific architectural components (Simultaneous modeling of multiple variables. %Tying to explainability. ) -> (2) 
% use physics to build architectures that replace existing architectures (e.g., graphs, recurrent units),
% (2) modularized/ neural networks, 
% -> (3) use physics to motivate the development of entirely new ML architectures, e.g., the use of diffusion processes in stable diffusion models?
% Structuring the embedding space of ML models using scientific knowledge. Explainability. Discovery of meaningful intermediates.
% }

\begin{figure}[!h]
\begin{center}
    % \vspace{-24pt}
	\includegraphics[scale=0.37]{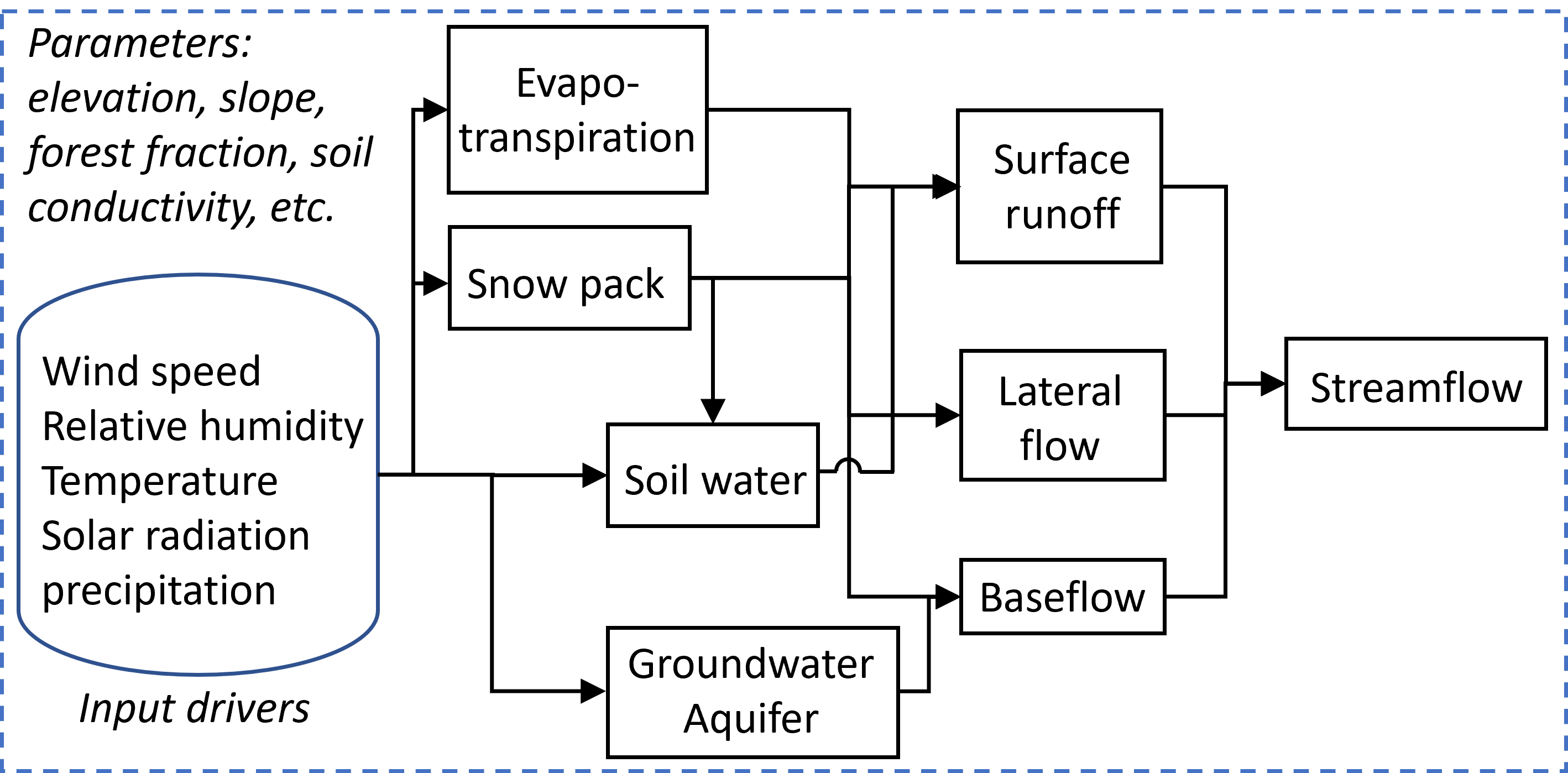}
\end{center}
    \vspace{-10pt}
	\caption{The hierarchical physical variables simulated in the hydrological systems. % by the SWAT model. 
	The variables shown on each module is the output of the module. %The streamflow on the right is the final target variables.
	}
	\label{fig:hydrology_flow}
\end{figure}

ML models are essentially parameterized architectures that need to be trained using observational data. The internal structures of ML models often involve common components such as convolutional layers, recurrent layers, and graph layers for capturing complex spatio-temporal data dependencies. However, these data-driven components are prone to overfitting when the available observations are limited or not representative of the data distribution in larger regions or different time periods. There is a general interest in integrating scientific knowledge into specific components in ML architectures~\cite{jia2019physics,anonymous2023climode,jia2021physics_pgrgrn,bao2021partial,dugdale2017river,lienen2022learning,anonymous2023airphynet}. These methods can be broadly grouped into three categories: (1) incorporating knowledge to modify components in existing deep learning models (e.g., convolutional and recurrent layers, graph structures), (2) decomposing ML models guided by the interactions of different processes, and (3) encoding physical properties (e.g., invariance or equivariance) into the architecture.

Prior work has explored different ways to enhance the deep learning architecture by integrating scientific knowledge into certain modeling components~\cite{lienen2022learning,bao2022physics,bao2021partial,jia2019physics,jia2021physics_pgrgrn,daw2020physics,anonymous2023airphynet}. 
In the context of water temperature prediction in freshwater ecosystems, prior works have incorporated the energy conservation law in modeling the hidden information being transferred across time in the RNN structure~\cite{jia2019physics} and across stream segments in the graph convolutional process~\cite{jia2021physics_pgrgrn}.  
Other works have explored directly using known PDEs to inform certain model components. For example, Bao et al.~\cite{bao2021partial} built dynamic graph structures across stream segments based on the heat transfer PDE~\cite{dugdale2017river}.  Lienen et al.~\cite{lienen2022learning} developed a graph neural network by incorporating physics knowledge on the unknown PDE to improve sea surface temperature and gas flow predictions. Airphynet~\cite{hettige2024airphynet} also utilized multiple graphs to capture the diffusion and advection effect in the air pollutant concentration. In the work by Daw et al. \cite{daw2020physics}, a physics-guided architecture of LSTM models (PGA-LSTM) was developed to explicitly encode the density-depth physics in the connections among LSTM nodes for the problem of lake water temperature modeling. PGA-LSTM was shown to produce meaningful uncertainty estimates of water temperature profiles across depth in lakes, where every realization of temperature profile generated by the model was guaranteed to respect density-depth physics.

% 1. recurrent layer (PGRNN), Graph layer (bao2021pde)

Another thread of research is in decomposing the model architecture into multiple modules responsible for unique physical processes, with interactions among processes %at different spatial and temporal scales 
captured by module dependencies.  
%modifying the overall model architecture to capture interactions amongst different physical processes  at different spatial and temporal scales. 
For example, 
the agroecosystem consists of diverse types of physical processes, e.g., weather, soil conditions, plant, and respiration, 
which jointly form the cycling of energy, water, and carbon. These processes are  simulated by existing physics-based models, %crop models 
%~\cite{grant2010changes,zhou2021quantifying,hipsey_general_2019,arnold2012swat,markstrom2012p2s}, % through a series of mathematical equations.
which  
are often built with a modular structure. Fig.~\ref{fig:hydrology_flow} shows a general modular structure of many hydrology models %, e.g. SWAT~\cite{arnold2012swat}, 
for simulating streamflow.  Here each module is responsible for converting input variables to specific output variables based on known or modeled physical processes, and these variables can be either observable or unobservable.    Each module %converts input variables to output variables through 
is implemented as a series of mathematical computations, which can be either explicit (e.g., following mathematical equations) or implicit (e.g., solving a differential equation). 
%Even though some equations used in physics-based models are only approximations of reality, the knowledge embodied in these models can be used to enhance the capacity of ML models and reduce the hypothesis space in the optimization process. %In the following, we will describe  
Conventional ML largely ignores the modular structure and try to directly create a mapping between input drivers (e.g., weather) and the response variable (e.g., streamflow). 
To fully capture the dynamics of variables and their interactions in scientific systems,  %This motivates leveraging 
%we plan to incorporate the knowledge from physics-based models to enhance the ML model.
%we plan to develop 
previous methods developed modularized
NN architectures, which are consistent with the processes defined in physics-based models~\cite{liu2022kgml,kraft2022towards,liu2024knowledge,he2023physics}.    
This entails ascribing physical meanings to the output 
of each NN module %intermediate hidden variables 
while also maintaining the interrelationships amongst these modules %physical variables 
as defined in existing physics-based models. 
This architecture decomposes the modeling of different variables %that may evolve at different scales 
and thus %also 
offers the flexibility for selecting different NN structures (e.g., recurrent and convolutional networks) for each module that best fit the nature of each variable that may evolve and interact at different scales. %specific prediction task. 
For example, Liu et al.~\cite{liu2022kgml} created a hierarchical NN structure using multiple GRUs for predicting different intermediate physical variables in agroecosystems, such as CO$_2$ fluxes and soil NH$_4^+$. 
Feng and Shen et al.~\cite{feng2022differentiable,shen2023differentiable} further extended this idea by replacing certain process-based modules %in the physics-based model 
by NN structures  %. They also 
while implementing other time-discrete processes directly on PyTorch to allow automatic differentiation. They also 
model the spatial variation of physical parameters (in the process-based model) %over different regions %. In particular, they built a 
using a separate NN model (e.g., LSTM)~\cite{tsai2021calibration}, which takes the forcing data and attributes to predict the static parameters or dynamic parameters of the process-based model, and feeds them to the main predictive model.%, which are then fed to the main predictive model. 

% \textcolor{red}{Todo: add equivariance structures (e.g., steerable CNN, EGNN) for material science/moledule modeling. }

In addition, the ML architecture can be designed to explicitly capture desired physical properties, e.g., the equivariance property in a dynamical particle system. While the equivariance can also be partially achieved through data augmentation, directly encoding it in model architecture can reduce the number of model parameters, and thus mitigate overfitting and improve the generalizability. Most existing equivariance models (e.g., G-CNN~\cite{cohen2016group}, Steerable CNN~\cite{cohen2016steerable}) are developed by extending the translation equivariance of the standard CNN model to group equivariance. 
%Steerable CNN
The EGNN model~\cite{satorras2021n} further extends this to higher dimensional space on a graph structure. It is also noteworthy that the development of many commonly used ML architectures was initially inspired by important physical properties. For example, the convolutional layers in the standard CNN are designed to preserve translational equivariance in image recognition.

\subsection{Knowledge-guided Pre-training}
% \textcolor{red}{Connections between physics model parameters and ML parameters. The source of knowledge to be used in model pre-training (e.g., simulated data, sample-wise similarities, causal relations amongst different physical variables)
% }
Overparametrized ML models are prone to overfitting due to the large parameter space and the instability inherent in their optimization process. Scientific knowledge can be leveraged to initialize the model parameters so less data is needed to refine the model parameters to learn generalizable data patterns. Existing KGML pre-training methods can be conducted by utilizing simulated data or self-defined proxy learning tasks (self-supervised learning), as discussed in the following.  

\textbf{Use of Simulated data:} 
Simulated data generated by process-based models (even generic uncalibrated models) can be used to pre-train ML models. Such simulated data encode fundamental physical principles governing the complex systems, and thus 
can provide valuable information in training ML models, especially when observations are sparse.  
Use of such pre-training is one of the easiest ways of leveraging the knowledge about scientific principles governing complex environmental systems into ML~\cite{xu2023deep, jia2019physics, jia2021physics}. Jia et al.~\cite{jia2019physics} first utilized simulated data from a generic process-based model to pre-train the ML model and demonstrated the improvement in the performance under data-sparse and out-of-sample scenarios.

% Prior work has explored using {simulated data} for model pre-traning. Generally, the simulated data are generated by the physics-based models, which represent the foundational principles governing the complex system of environmental science problems. Prior work has shown the promise of pre-training ML models using simulated data, especially when observations are sparse~\cite{xu2023deep, jia2019physics, jia2021physics}. 

% imperfect simulations /parameters or other factors not considered
One limitation of this approach is that the simulated data are often biased due to the parameterization used in physics-based models, which makes the pre-training less helpful when we adapt (or fine-tune) the model to the target system.  %leading to potential negative transfer after pre-training. 
In particular, physics-based models simulate different physical processes with preset parameterizations, e.g., %  simulated data is created by systematically varying the parameters within a reasonable range defined by the physics-based models. 
%For instance, one may manipulate the parameters representing hydrological conditions, such as precipitation levels, 
the setting of soil moisture and watershed characteristics when simulating streamflow and water temperature, but these preset parameter values can be very different in the target ecosystems.  %, are manipulated based on the physics-based models. 
Several approaches have been developed to address this issue~\cite{jia2021simlr, chen2023physics, jia2022modeling, chen2023physics_stream}. For example, Jia et al.~\cite{jia2021simlr} proposed a meta-learning algorithm to initialize an ML model using multiple sets of simulated data created by varying the parameterizations of the physics-based model.  %within a  range according to the prior knowledge about the target system
This process results in a diverse dataset that closely emulates real-world hydrological data, %. %The aim is to ensure that the simulated data accurately reflects the expected variations and behaviors of the actual hydrological system, 
allowing the ML model to learn and generalize effectively to real-world scenarios. McCabe et al.~\cite{mccabe2023multiple} also explored embedding data from multiple physical systems into a shared representation space so as to create a generalizable pre-trained model.  In another work, Chen et al.~\cite{chen2023physics} utilized multiple physical equations to simulate baseflow in river basins, and selectively used simulated values to regularize the ML model in its initialization and adaptation phases. 

Another issue with simulated data is that they are often generated independently by physics-based models without considering the effect of other real-world factors, e.g., human infrastructures. 
For example, most physics-based stream models do not explicitly capture the effect of dams,  reservoirs, and transportation systems, which introduces bias to the simulated streamflow and water quality measures. 
Such biased simulated data can be less helpful in initializing ML models as they present only physical processes that are in ideal scenarios but could behave very differently from true observations. 
One interesting direction is to combine simulated datasets independently created over different systems and use data-driven approaches to model their interactions. 
Recently, there was a data release from the U.S. Geological Survey on stream temperature simulations that combines two types of physics-based models, the SNTemp model~\cite{bartholow2010stream} for streams and the GLM model for reservoirs~\cite{hipsey2019general}. Such a combination is through a linear function between each stream and its upstream reservoir, and the weight of the reservoir is exponentially decayed over a longer distance.  Jia et al.~\cite{jia2023physics} further extended this idea by using a graph neural network to combine the simulations on different types of nodes, and then use the obtained composite simulations to pre-train the graph model for predicting stream water temperature.

\textbf{Self-supervised learning: } %Sample-wise similarities and predictive methods 
% include molecule paper
In the field of self-supervised learning (SSL), deep neural networks are trained to learn useful feature representations using pseudo labels created from pre-defined pretext tasks, e.g., colorization~\cite{larsson2017colorization} %~\cite{zhang2016colorful,larsson2017colorization} 
and inpainting~\cite{pathak2016context} in computer vision.  
These pretext tasks are designed in a way such that solving these pretext tasks requires the extraction of complex data patterns needed for the target prediction task. %The domain knowledge can be leveraged in the SSL process for pre-training ML models. 
For the self-supervised methods 
to be effective in environmental science problems, one promising direction is to create pretext tasks %do not 
that can reflect knowledge about underlying physical processes. 
In general, the self-supervised learning paradigm can be broadly categorized as sample-wise similarity learning and predictive learning, described in the following.

The core concept of sample-wise similarity learning involves learning representations for individual data samples, aiming to precisely reflect the inherent similarities between them~\cite{yan2023adaptive,wan2023molecules}. This approach primarily concentrates on deciphering the inherent relationships within the data, considering each data point in relation to every other point in the sample space. 
Specifically, a similarity is estimated between each pair of samples based on their characteristics in scientific systems, e.g., depth and surface area in lake systems~\cite{willard2021predicting,chen2023physics, chen2022physics}. %~\cite{}to encode the underlying structures and relationships in the data. 
Next, 
ML models are pre-trained to map data samples to a representation space, in which  % this methodology aim to position 
similar samples are pulled closer and dissimilar ones are pushed further apart~\cite{ghosh2022robust,wei2021large}. % in the learned feature space, achieved through various methods 
This can be commonly achieved by contrastive learning~\cite{le2020contrastive}. 
% For instance, in analyzing baseflows within different river basins, differentiating the similarities and contrasts between basins under diverse hydrological conditions can aid in decoding the underlying hydrological processes, improving the prediction of baseflow dynamics. 
For example, Ghosh et al.~\cite{ghosh2022robust} adopted this in streamflow prediction. They developed a representation learning method to embed basin-specific characteristics, which is trained via contrastive learning using positive (similar) sample pairs from the same basin and negative sample pairs from different basins.

As an alternative method, predictive SSL learning aims to pre-train ML models by predicting certain state variables in the target system. The idea is to enforce the learning of key state variables related to the target variables.  For example, an ML model for crop yield prediction can be pre-trained to predict variables related to its carbon and water cycles~\cite{licheng2022estimating}. The training can be conducted using simulated values or proxy values for these state variables. 
Prior works further extended this idea to pre-training a knowledge-guided ML architecture, which contains multiple ML modules to represent different processes~\cite{liu2022kgml}. They leveraged simulated data for intermediate modular outputs to pre-train different ML modules.

% \textbf{Leveraging relationship amongst different variables.} 

% are identified based on domain knowledge, experimental data, or causal discovery algorithms. Once causal relations are identified, incorporate this knowledge into the ML models as priors or constraints to enforce physically plausible representations and predictions. In hydrology, by embedding detailed causal relationships—such as the impacts of land use and land cover (LULC) changes on runoff generation—and pre-training ML models based on these relationships, models are tailored to mirror established hydrological principles. This sets the stage for learning the subsequent relationships in the causal chain, thereby enhancing our capacity to understand and predict complex hydrological phenomena, like river water quality, under varying land-use scenarios.

% how to do fine-tuning/probing 

% After selecting appropriate machine learning models and defining their architectures, specific parameters within these machine learning models are initialized using information derived from the physics model. These initialized parameters remain fixed and unaltered during the initial training phase, effectively preserving the foundational knowledge rooted in the physics-based model. Subsequently, the model undergoes fine-tuning on the simulated data, enabling it to adapt and optimize its parameters for the specific environmental problem. \textcolor{red}{Other probing approaches?}

\textbf{The Challenge of Catastrophic forgetting: } 
The pre-training enables the ML model to learn the representation of physics from simulated data. Next, the model can be fine-tuned with true observations from the target system~\cite{sharma2023knowledge}. 
One potential issue is that the fine-tuning could distort the representation learned from pre-training, leading to degraded generalizability~\cite{kumar2022fine}. There is an opportunity to explore different fine-tuning techniques, e.g., tuning a subset of parameters at a time while fixing other parameters~\cite{jia2021simlr}. The objective is to preserve the generalizable patterns extracted from simulated data when adapting the pre-trained model to the target system.

\section{Use Cases in Environmental Sciences}
\label{sec:use-cases}

\begin{figure}
\centering
\includegraphics[width=\linewidth]{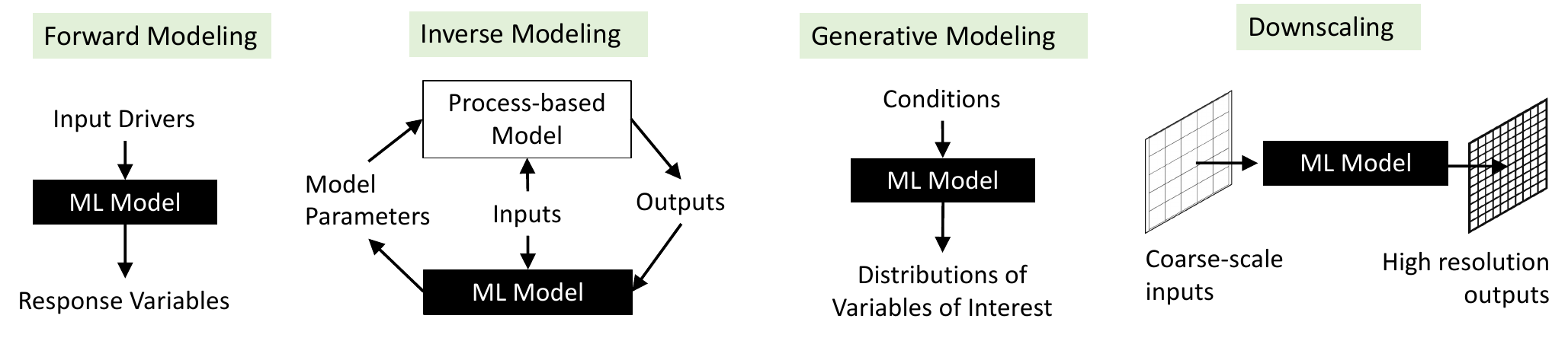}
\caption{Schematic diagrams of four primary use-cases of KGML in scientific disciplines
}
\label{fig:kgml_usecases}
\end{figure}

Figure \ref{fig:kgml_usecases} illustrates many prominent use-cases for using KGML methodologies in scientific problems. First, KGML methods can be used to learn the forward mapping from input drivers to response variables {better} than state-of-the-art (or sometimes, non-existent) science-based forward models in terms of accuracy and/or computational efficiency. A second use-case is to learn the inverse mapping from observations of inputs and outputs to the parameters of science-based models using ML algorithms, which can help provide useful information of the system states and aid in calibrating physics-based models. %For example, in the problem of seismic imaging, ML methods have been developed to learn the inverse mapping from observations of waveform amplitudes to the velocity profile of wave propagation through different layers of the Earth's subsurface \cite{jin2021unsupervised}. A key motivation behind inferring the parameters of science-based models is to recover critical  information about the state of scientific processes that are difficult to observe directly. Another motivation is to identify useful settings of model parameters that when fed inside science-based forward models result in simulated outputs that best match with observations, referred to as the problem of model calibration. Note that inverse modeling can also be performed with partial knowledge of the underlying system without explicitly using a science-bases forward model, as explored in a recent work in the domain of hydrology \cite{ghosh2022robust}. Another recent line of work in KGML for inverse modeling includes the framework of differentiable parameter learning (dPL) \cite{tsai2021calibration} for geoscience applications, where deep learning methods are used to infer the parameters of differentiable physics-based models using gradient descent algorithms leveraging automatic differentiation of the models. 
A third use-case of KGML in scientific domains is to use generative modeling approaches to create digital twins of scientific systems, where ML methods are used to generate synthetic distributions of variables of interest (e.g., spatio-temporal variations of Earth's surface temperature), optionally conditioned on a few parameters. This is useful for the purpose of data generation, which can then be fed into downstream tasks of forward modeling. 
A fourth use-case of KGML in scientific domains is to convert coarse-scale information of physical variables to higher resolutions in space and/or time, referred to as the problem of downscaling (also referred to as the problem of super-resolution in the field of computer vision). This can help reduce the expensive computational cost of creating simulations at fine scales. %Note that the coarse-scale inputs can either be obtained through observations or through simulations of low-fidelity models operating at coarser scales. Also, we generally require paired examples of coarse and high-resolution data to train ML models for the purpose of downscaling. An exception to this general rule is the framework of Fourier neural operators \cite{li2020fourier} that are able to perform zero-shot downscaling of solutions of PDEs without using any data (either simulated or observed) at higher resolutions.

\subsection{Forward Modeling}
% can be done even without knowledge of PDEs.

Process-based models consist of a series of mathematical equations to simulate the forward mapping of underlying processes from input drivers to response or target variables. Running these models often requires substantial computational costs due to the need to solve internal equations (e.g., complex PDEs). This poses a major impediment for simulating scientific processes at fine spatial resolutions and high time frequency required in many real-world settings. Moreover, existing process-based models often use parameterizations and approximations due to incomplete knowledge or excessive complexity in modeling certain processes, leading to degraded predictive performance. ML approaches are increasingly being utilized to build forward models, but standard ML approaches are not designed for capturing complex dynamics and interactions of scientific processes, especially when training data are limited. 

KGML methods can be used in the problem of forward modeling in two scenarios: 
(1) surrogate modeling: when target variables can be assumed to be perfectly described by process-based models, e.g., simulated data generated by solving PDEs, KGML methods can be utilized to build computationally efficient surrogate models, and  
(2) improved forward modeling: when we have observational data for target variables and process-based models are known to be imperfect (or approximate) and incomplete representations of reality, KGML methods can be used to improve both computational efficiency and predictive accuracy of forward modeling. We describe both these scenarios in the following.

% Two categories: (1) surrogate model for numerical PDE solver (with or without the knowledge of PDEs), and (2) surrogate model for other complex process-based models

\subsubsection{Surrogate Modeling}

\textbf{Solving PDEs: }
Traditional numerical solvers for PDEs often use the Finite Elements Method or the Finite Difference Method \cite{jacob2007first}, and can be prohibitively expensive for many simulation tasks, e.g., simulating turbulent flow with a large Reynolds number. %For example, the direct numerical simulation (DNS) of incompressible turbulent flow with  Reynolds number $\text{Re} = {\cal O} (10^5)$ within  a domain of size of ${\cal O}\left[(100 \ell)^3\right]$  requires about {\it a century} of CPU time on a 1 teraflop computer. 
Recently, neural network (NN) models have been used to approximate the solution of PDEs through the neural network (NN) forward process, which significantly reduces the computational cost compared to conventional numerical solvers. NN-based PDE solvers have shown promise in approximating PDE solutions with reasonable accuracy in a number of applications. Some popular approaches include physics-informed neural networks (PINN)~\cite{karniadakis2021physics} that are designed to solve a single instance of a PDE, and neural operator learning methods that can learn a family of PDEs including the Fourier Neural Operator (FNO)~\cite{li2020fourier} and DeepONet~\cite{lu2019deeponet}.  Operator learning methods such as FNO and DeepONet can be further enhanced by adding physical constraints in the training process~\cite{li2021physics,wang2021learning}.

One area of the recent focus in the problem of solving PDE is in using  {FNO}   to approximate the PDE solutions through a transformation in a Fourier space. The intuition is to approximate the Green functions by kernels, which are parameterized by neural networks in the Fourier space.   %by leveraging the Fourier transformation and neural networks. 
Specifically, the FNO approach initially transforms the input function, representing the initial or boundary condition of a PDE, from the spatial domain to the frequency domain using the Fourier transformation, which allows the representation of the function to capture the global information effectively. A neural network then learns the mapping between the Fourier coefficients of the input and output representation in the frequency domain. After this learning process, the inverse Fourier transform is applied to convert the learned representation back to the spatial domain, yielding the approximated solution to the PDE. Based on such network design, FNO is able to combine the global information of the entire field embedded through the Fourier transformation and the expressive power of neural networks, enabling the learning and approximation of high-dimensional and complex PDE operators directly from data. 
An additional benefit of FNO is its ability to create the PDE solution at arbitrary spatial resolutions. This has the potential to achieve zero-shot or few-shot downscaling in complex dynamical systems.  Several variants of FNO have already  emerged~\cite{kovachki2021neural, tran2021factorized, zhao2022incremental, lutjens2022multiscale} and have also been used to address the scientific challenges in environmental science~\cite{li2022fourier, jiang2023efficient, yang2023fourier}. The PICL approach~\cite{lorsung2024picl} was introduced to further extend FNO for simultaneously modeling multiple governing PDEs by using contrastive learning to enforce the distance amongst samples from different PDE systems.  
Furthermore, Lippe et al.~\cite{lippe2023pde} introduced the diffusion model~\cite{wijmans1995solution} as  PDE-Refiner into FNO for more accurate modeling of PDE dynamics via a multi-step refinement process.

\textbf{Surrogate Modeling for Other Systems:}
% Surrogate modeling in environmental science is a technique that simplifies complex process-based models, making them more computationally efficient. 
Existing process-based models rely on intensive  computations to simulate complex physical processes. 
%These complex models, which accurately simulate environmental systems and processes, are often highly detailed and computationally demanding. 
Surrogate modeling addresses this challenge by creating data-driven models that approximate the behavior of physical models and predict the outputs of these complex systems with much less computational effort. This makes them extremely valuable in scenarios where rapid decision-making is crucial, such as emergency response planning, environmental impact assessments, or policy development~\cite{razavi2012review, fan2020optimal, liu2021study, zhang2022prediction, bocquet2023surrogate, zhu2022building, himes2022accurate, zahura2022predicting}.
Surrogate models are typically developed using advanced machine learning tools ~\cite{bocquet2023surrogate, zhu2022building, himes2022accurate, zahura2022predicting,tayal2023koopman,wang2023high} (e.g., Koopman operator~\cite{brunton2017koopman}) or statistical methods~\cite{furtney2022surrogate}, and are trained on simulated data generated by complex physical models. 
To fully capture %, enabling them to capture 
the essential patterns and relationships inherent in environmental processes, %the surrogate models can also be enhanced using physical constraints~\cite{} and 
scientific knowledge can also be leveraged to enhance the ML-based surrogate models. 
% In addition to adding sc constraints or loss functions, another opportunity in KGML is to have the ML architecture encode the causal dependencies amongst different components in the physical model~\cite{khandelwal2020physics}. 

% Once trained, the surrogate models can quickly provide analyses, making them extremely valuable in scenarios where rapid decision-making is crucial, such as emergency response planning, environmental impact assessments, or policy development~\cite{razavi2012review, fan2020optimal, liu2021study, zhang2022prediction, bocquet2023surrogate, zhu2022building, himes2022accurate, zahura2022predicting}. In the context of river flow prediction, a surrogate model is used to forecast river flows during flood events. Traditional hydrological models, complex and time-consuming, typically consider various parameters such as rainfall intensity, soil moisture, and landscape features to simulate river behavior. However, a surrogate model, trained on simulated data and grounded in key hydrological principles, can swiftly estimate river flow levels based on current weather forecasts and land conditions. This capability for rapid prediction is invaluable to emergency response teams and urban planners, enabling them to make timely decisions for flood preparedness and mitigation~\cite{meert2018surrogate, yosefipoor2022adaptive}.

\subsubsection{Improved Forward Modeling}

Utilizing scientific knowledge can further enhance the ability of process-based models to better align with ground-truth observations. 
Building upon the traditional approach of residual modeling where ML or statistical methods are used to learn the residual of a process-based model w.r.t. ground-truth data~\cite{thompson1994modeling}, researchers have investigated the combination of simulated and observational data through a pretrain-finetune pipeline~\cite{liu2022kgml, han2023deeporyza, licheng2023knowledge}. This method leverages the knowledge embedded in comprehensive simulated data while further improving the prediction on observational data through parameter refinement. %allows for the development of robust and reliable models. 
Another KGML approach for improved forward modeling is to incorporate domain knowledge into the ML model's architecture~\cite{liu2022kgml}. This integration ensures that the model adheres to established scientific principles, while the model parameters can be further adjusted to match true observations. 
Advancements also be made through feature selection, which utilizes domain knowledge to select or extract the most relevant features for the target task~\cite{xu2024knowledge}. This enhances the model's ability to focus on the most informative aspects of the data, thereby improving its predictive performance.  
Moreover, the ML model can be constrained by known physical relationships. 
%ensuring physical consistency in the model’s predictions is crucial. 
This can be achieved by either adding physical laws as constraints during the model training process~\cite{yang2023flexible, kashinath2021physics} or adjusting model outputs in the post-processing phase to maintain physical consistency~\cite{chen2024reconstructing}.

\subsection{Inverse Modeling}

Standard forward models aim to create a mapping $f$ from input drivers $x_t$ to target variables $y_t$. 
The mapping $x_t\rightarrow y_t$ can be different for every system (e.g., different basins or farmlands) due to the variation of physical characteristics (e.g., groundwater, hydraulic conductivity, soil, and vegetation conditions)~\cite{ghosh2023entity}. Hence, forward models often need to include physical characteristics $c$ as additional input, i.e., $y_t = f(x_t, c)$. 
Most physical characteristics $c$ are considered static or evolving slowly. 
Due to the %difficult to measure due to  the 
technical difficulties or prohibitively expensive cost in measuring  these characteristics,  %(e.g., groundwater residence time, soil properties), 
they are often modeled as physical parameters and  need to be calibrated %in the physics-based model 
using real observations $\{x,y\}$. 
Traditional calibration of physics-based models commonly %rely on 
uses grid search or Bayesian approaches to %search over the space of 
find combinations of parameter values %to achieve 
leading to the best match with observations or measurements, which can be highly time-consuming  
and require extensive domain expertise to select the range for each variable~\cite{hill2006effective,makela2000process,fatichi2016overview}. 
This process %can be very time consuming and also 
can also degrade the predictive performance using the obtained parameters due to the limited computational resources  available to explore all the possible parameter combinations, especially in scenarios involving complex interactions among %a %large number of  
variables and parameters.

Advances in ML and deep learning have brought opportunities to 
estimate physical characteristics directly using data-driven approaches. Given some available data samples $\{x_t, y_t\}$, the characteristics (or parameters) $c$ can be estimated by learning the  
inverse mapping $c = g(x_t,y_t)$. 
A key motivation behind inferring the parameters of process-based models is to recover critical information about the state of scientific processes that are difficult to observe directly. Another motivation is to identify useful settings of model parameters that when fed inside process-based forward models result in simulated outputs that best match with observations, referred to as the problem of model calibration.

Building upon traditional inverse modeling approaches based on regularized regression~\cite{engl1996regularization}, researchers are now building  ML models for the inverse modeling of characteristics in hydrology~\cite{ghorbanidehno2020recent}, photonics~\cite{pilozzi2018machine}, land surface temperature~\cite{wang2024simfair}, %land surface properties~\cite{dawson1992inversion}, 
among many others. 
For example, in the problem of seismic imaging, ML methods have been developed to learn the inverse mapping from observations of waveform amplitudes to the velocity profile of wave propagation through different layers of the Earth's subsurface \cite{jin2021unsupervised}. 
%These methods still need to address the sparse 
%several issues. First, to complement the sparsely observed target variables $y_t$, the inverse model.
%needs to utilize the abundant input drivers $\{x_t\}$  (e.g., meteorological drivers, which are much easier to obtain) for locations and time periods in which target variables $y_t$ are not measured. 
Some existing works on discovering PDEs %through the estimation of coefficients 
can also be viewed as special cases of data-driven inverse modeling through the estimation of PDE coefficients from available simulations~\cite{raissi2018deep,rudy2017data}. 
Note that inverse modeling can also be performed with partial knowledge of the underlying system without explicitly using a process-based forward model, as explored in a recent work in the domain of hydrology \cite{ghosh2022robust}. Another recent line of work in KGML for inverse modeling includes the framework of differentiable parameter learning (dPL) \cite{tsai2021calibration} for geoscience applications, where deep learning methods are used to infer the parameters of differentiable physics-based models using gradient descent algorithms leveraging automatic differentiation of the models.

Due to the intrinsic complexity of the inverse mapping, training inverse models can be challenging for some systems with sparse observations. Prior work has applied physical constraints to NN-based inverse models by regularizing the predicted latent state variables from observations~\cite{ishitsuka2023physics}. 
Another promising direction is to build an inverse loop for the forward model (e.g., using an invertible NN structure~\cite{tayal2022invertibility}), which allows natural incorporation of scientific knowledge in the forward process.  %while building an to guide the training process of the inverse model. 

% \textcolor{red}{connections to PDE discovery.}

\subsection{Generative Modeling}

Generative models have found great success in computer vision~\cite{wang2021generative,croitoru2023diffusion} and natural language processing~\cite{iqbal2022survey}. Some popular models include generative adversarial networks~\cite{goodfellow2014generative}, variational autoencoder~\cite{kingma2013auto}, normalizing flow-based modes~\cite{kobyzev2020normalizing}, and diffusion models~\cite{ho2020denoising}. These models aim to capture the underlying data distribution and then create synthetic samples that look similar to the real observations.     
Given their success in generating images, speech, and text data, the expectation is rising for using these models to generate virtual physical simulations under specified conditions.

Despite the power of these models, they are often found to generate spurious data patterns or artifacts when directly applied to complex applications in fluid dynamics and environmental sciences, especially when training data are scarce (e.g., simulations at high resolutions). 
To address this challenge, prior work has explored regularizing the generative models by additional physical constraints, which have shown improvement in the quality of generated data in fluid dynamics~\cite{yang2019enforcing,wu2020enforcing,chen2021reconstructing}, material science~\cite{cang2018improving}, and climate science~\cite{gao2023prediff}. 
For example, conservation laws have been utilized to regularize the GAN-based for generating the simulation of turbulent flows~\cite{chen2021reconstructing}. 
Similarly, 
Gao et al. also include the anticipated precipitation intensity for regularizing the denoising process of the diffusion model, and show the promise of this method for precipitation nowcasting and the detection of extreme cases such as rainstorms and droughts~\cite{gao2023prediff}. 
The generative modeling approaches have also been used for refining the initial predictions. For example, 
Lipple et al.~\cite{lippe2023pde} found that data-driven PDE solvers (e.g., FNOs) often neglect components of the spatial frequency spectrum that have low amplitude. To overcome this issue, they developed PDE-Refiner, which employs diffusion models to refine the prediction by considering information from all frequency components at different amplitude levels.  %iterative refinement process using%drawing inspiration from recent breakthroughs in diffusion models. PDE-Refiner represents a new category of models that employs an iterative refinement process to achieve precise predictions across the entire frequency spectrum. This level of accuracy is attained through a modified Gaussian denoising step, which ensures that the network considers information from all frequency components equally at different amplitude levels~\cite{lippe2023pde}.

\subsection{Downscaling} 
% \textcolor{red}{How to preserve physics in data downscaling, especially in zero-shot or few-shot scenarios? Also the data can be either gridded (e.g., images) or irregularly structured (e.g., stream networks). }

In environmental science problems, complex physics-based models are used to capture the details of physical reality by incorporating diverse components that account for numerous processes at fine spatial or temporal scales. These models are often restricted by computational expenses and model complexity. Consequently, many models are operated at a coarser resolution than necessary to accurately represent the underlying physical phenomena. For instance, cloud-resolving models (CRMs) effectively capture boundary layer eddies and low clouds with sub-kilometer horizontal resolution~\cite{rasp2018deep}. However, running global climate models at such fine resolutions, even with anticipated advancements in computing power, remains unfeasible. 

Handling downscaling problems in environmental science using machine learning involves deriving fine-scale information from coarser-scale data. 
Note that the coarse-scale inputs can either be obtained through observations or through simulations of low-fidelity models operating at coarser scales. In particular, machine learning models have been widely used for automatically projecting the coarser-scale or lower-resolution environmental data into fine-scale or higher-resolution ones, which is especially popular in the domains of climate science, hydrology, and ecology~\cite{jebeile2021understanding, yeganeh2022machine, schneider2022machine, liu2020downscaling, guevara2019downscaling, sabzehee2023enhancing, willard2022integrating, bennett2021characterization, kashinath2021physics, park2022downscaling}. For example, Wang et al.~\cite{wang2021deep} adopted super-resolution methods to generate higher-resolution predictions (e.g. temperature, precipitation, etc)  in different locations and times at the local scale from coarse spatial resolutions. The authors further extended this work by transferring % through transfer learning, 
the trained model in one region %was directly applied 
to downscale the precipitation in another region under a different environment. 

In many environmental science problems, building data-driven downscaling methods is challenging due to the often limited availability of fine-scale observational data. Few-shot and zero-shot learning could be potential solutions to such problems. Few-shot learning~\cite{wei2022mapping} and zero-short learning~\cite{jiang2023efficient, yang2023fourier, harilal2021augmented} can leverage models trained from data-sufficient regions
and adapt them to regions that are poorly observed or completely unobserved after moderate fine-tuning. %, fine-tuning them using sparse local observations
%~\cite{wei2022mapping}. 
%Meanwhile, zero-shot learning leverages attribute-based embeddings from known regions to predict outcomes in areas without direct observations~\cite{jiang2023efficient, yang2023fourier, harilal2021augmented}. 
%While these approaches are nascent in environmental science, they 
%While thold 
Despite the promise of these approaches, using these approaches for improving scientific data downscaling remains largely under-explored. 
%Also, we generally require paired examples of coarse and high-resolution data to train ML models for the purpose of downscaling. An exception to this general rule is the framework of Fourier neural operators \cite{li2020fourier} that are able to perform zero-shot downscaling of solutions of PDEs without using any data (either simulated or observed) at higher resolutions. 
Recently, FNO has shown encouraging results in performing zero-shot downscaling of PDE solutions to arbitrary specified resolutions~\cite{li2020fourier}.  %in data-scarce scenarios.
Another limitation of existing data-driven downscaling approaches is their ability to harness irregularly structured data. %In another aspect, d
Existing downscaling techniques in environmental science are often applied to gridded data, such as remote sensing imagery. %, but are not suitable for  
%or 
%irregularly structured data, such as river stream networks~\cite{}. 
There are plenty of machine learning models suitable for these kinds of data, e.g., %. For gridded data, 
Convolutional Neural Networks (CNNs) and Generative Adversarial Networks (GANs) are commonly used in super-resolution (SR) methods for refining coarse spatial datasets %, such as satellite imagery
~\cite{lambhate2020super, ballard2022contrastive}. 
More recently, researchers have started exploring downscaling approaches for 
%For 
irregularly structured data, such as river networks ~\cite{chen2021heterogeneous}. In prior work, a graph network model is used to represent and predict intricate network dynamics, while a long short-term memory (LSTM) model is used to capture flow sequences along river paths~\cite{sit2023spatial}. These methodologies hold promise for enhancing fine-scale environmental predictions from coarser datasets.

% \textcolor{red}{discussions on FNO.}

% \section{Emerging Opportunities in KGML Research}
% \label{sec:cross-cutting}

\section{Foundation models in environmental science}
\label{sec:cross-cutting}

Existing process-based and data-driven models are often built for specific target problems in environmental science. The coupling of multiple models is often non-trivial, making it difficult to capture the relationships or share the information across different problems.   For example, the prediction of water quality variables (e.g., water temperature, concentration of nutrients) and water quantity variables (e.g., streamflow) are often separately modeled even though they involve some common processes. 
Foundation models, which aim to establish a general paradigm to tackle a diverse set of tasks, have found immense success in natural language processing (NLP)~\cite{bommasani2021opportunities,zhao2023survey}. %which previously require separate models. 
% LLMs
Utilizing the textual understanding capabilities acquired from extensive text data in pre-training, these models have been shown to provide reasonable predictive performance on different tasks under zero-shot or few-shot settings.  
The intricate structure of the large-scale model also allows the information to be forwarded through numerous branches of routes in the model, facilitating the encoding of both shared and distinct information essential for various tasks.  This enables learning simultaneously over diverse tasks and then generating predictions for any task specified by a user.

Recently, researchers started exploring using foundation models (e.g., LLMs) in environmental applications~\cite{xie2023geo,mai2023opportunities,ai4science2023impact}. When used in environmental science applications,  foundation models provide new opportunities for performing predictions for different physical variables and at different scales, or over different sets of locations and different time periods. The expectation is also rising for using foundation models to understand the complex nature of the environment from limited observation data. 
Additionally, prior works have  shown that LLMs are effective in harnessing missing or inconsistent input data (e.g., weather data, soil properties, and other modeled variables) across space and assimilating new observations in environmental applications~\cite{luo2023free}. 

% Luo et al. also demonstrate that LLMs can be used to harness inconsistent input features and assimilate newly collected observations in predicting stream water temperature. 

Despite the promise, several major challenges could prevent existing foundation models from reaching the same level of success as they did in the text and vision domains. Environmental data are often not included in training existing foundation models. This omission can lead to a knowledge gap when the model is directly deployed for query tasks involving complex environmental processes. Although some open-sourced foundation models can be fine-tuned toward the downstream task, the size of observation data in many environmental applications is often far smaller than what is needed for tuning large-scale models.    Moreover, existing foundation models remain limited in interpreting model predictions from a scientific perspective, e.g., the simulation of intermediate processes that lead to the dynamics of final predicted variables. 
Third, running large-scale foundation models requires high computational cost and financial cost, which makes them unsuitable for prediction over long periods and large regions. To address these challenges, one promising direction is to  
better leverage scientific knowledge for enhancing the foundation models, e.g., through prompting, reasoning, and tuning,  to improve their generalizability and interpretability. New research is also urgently needed for selectively running foundation models and existing data-driven or KGML models in a hybrid manner to reduce the cost.

Alternatively, researchers also investigated building foundation models directly on environmental applications without referring to existing foundation models in vision and text domains.  For example, the ClimaX model~\cite{nguyen2023climax} has been built using climate datasets and a Transformer-based architecture. It was reported to achieve good predictive performance and generalizability on different climate and weather tasks.  
Several previous studies in scientific modeling can be considered as foundation models.  
For example, FNO~\cite{li2022fourier} has been shown to be able to perform PDE solving tasks for PDEs with different settings (e.g., different initial conditions). FNO and implicit neural  representation~\cite{chen2022videoinr} can %also be trained on a specific spatial resolution and then 
be used to downsample data to an arbitrary resolution after they are trained on a specific spatial resolution. 
Existing works on building heterogeneity-aware models over a large number of different locations can also be viewed as simplified foundation models that are specifically designed for spatial generalization~\cite{xie2022statistically,li2022regionalization,kratzert2019towards}. For example, Xie et al.~\cite{xie2022statistically} create a general pipeline for transforming deep learning models to predict land covers for locations with statistically different distributions. These models could be further extended to address the problems of different spatial scales.

% transfer learning
Another thread in building foundation models is based on the concept of transfer learning~\cite{zhuang2020comprehensive}. By using this approach, a model is first trained on a broad dataset that encompasses multiple scientific topics~\cite{lacoste2023geo}. This initial training allows the model to learn general trends and behaviors. After this broad training, the same model can be fine-tuned or adjusted on a more specific dataset related to a particular environmental task~\cite{nguyen2023climax}. Consider an example in hydrological problems. A foundation model trained using broad environmental features like climate data, satellite imagery, and general topography could be fine-tuned using river basin-specific data. This specialized model can predict river flow dynamics or water quality, optimizing its performance by leveraging both the broad knowledge from the global pre-training and the specific nuances from the fine-tuning phase. However, several challenges arise in this approach. Data in different scientific problems often vary in scale, resolution, and quality, making the integration a complex procedure. Moreover, the intricate interactions in environmental systems, like hydrology with its dependencies on soil, plants, and weather, make it challenging to capture all nuances.

\section{Concluding Remarks and Future Directions}
\label{sec:conclusions}
% \textcolor{red}{Move to Abstract:} This paper presents an overview of the problem of scientific modeling in the context of environmental applications, and discusses the complementary strengths and weaknesses of ML methods for scientific modeling in comparison to process-based models. It also provides an introduction to the current state of research in the emerging field of scientific knowledge-guided machine learning (KGML) that aims to use both scientific knowledge and data in ML frameworks to achieve better generalizability, scientific consistency, and explainability of results. We discuss different facets of KGML research in terms of the type of scientific knowledge used, the form of knowledge-ML integration explored, and the method for incorporating scientific knowledge in ML. We also discuss some of the common categories of use cases in environmental sciences where KGML methods are being developed, using illustrative examples in each category.

This paper provided a summary of prior research in KGML and contextualized it using a three-dimensional view. KGML is indeed a rapidly growing field of research, as illustrated by the diversity of research methods and scientific use-cases that are being explored in a variety of application domains. There are also several directions for future work in KGML. First, along with serving the goal of better predictive accuracy, novel advances in KGML are needed to solve the end-goal of discovering new scientific knowledge from data. While there is a growing focus in mainstream ML on explainability, trustworthiness, and fairness of ML results, scientific problems may require entirely new definitions of explainability and related concepts. 
% This is because the nature of understanding in scientific problems is fundamentally different from current notions of explainability used in mainstream ML to attribute input regions associated with predictions. 
This is because in scientific applications, we need to go beyond learning input attributions to make sure that the patterns, insights, or rules discovered from data are consistent with existing scientific theories and are meaningful to domain scientists. 
We also need to go beyond modeling associations to understand the causality of scientific systems, e.g., using counterfactuals, in conjunction with existing scientific knowledge.
Another advance that is needed in the field of KGML is to design better tools for uncertainty quantification in scientific applications of ML. For example, we need to improve our understanding of the limits of large-scale ML models (e.g., Foundation models) and predict in advance when ML models will fail and by how much given the characteristics of a new dataset.

\section{Acknowledgements}
We thank Shengyu Chen for his  valuable contributions to summarizing the existing literature on KGML research methods.  
We also thank Runlong Yu and Yue Wan for their very helpful review on an earlier version of the manuscript. This work was supported in part by National Science Foundation (NSF) awards IIS-2239328, IIS-2313174, IIS-2239175, IIS-2147195, IIS-2316305, OAC-1934721, OAC-2203581, and DEB-2213550. This work was also supported by USDA National Institute of Food and Agriculture (NIFA) and the National Science Foundation (NSF) National AI Research Institutes Competitive Award \# 2023-67021-39829.

\bibliographystyle{plain}
\bibliography{bibs/anuj,bibs/ch1-ramki,bibs/shengyu}

\begin{thebibliography}{100}

\bibitem{ai4science2023impact}
Microsoft~Research AI4Science and Microsoft~Azure Quantum.
\newblock The impact of large language models on scientific discovery: a preliminary study using gpt-4.
\newblock {\em arXiv preprint arXiv:2311.07361}, 2023.

\bibitem{anonymous2023airphynet}
Anonymous.
\newblock Airphynet: Harnessing physics-guided neural networks for air quality prediction.
\newblock In {\em Submitted to The Twelfth International Conference on Learning Representations}, 2023.
\newblock under review.

\bibitem{anonymous2023climode}
Anonymous.
\newblock Clim{ODE}: Climate forecasting with physics-informed neural {ODE}s.
\newblock In {\em Submitted to The Twelfth International Conference on Learning Representations}, 2023.
\newblock under review.

\bibitem{arif2022computational}
Muhammad~Shoaib Arif, Kamaleldin Abodayeh, and Yasir Nawaz.
\newblock A computational approach to a mathematical model of climate change using heat sources and diffusion.
\newblock {\em Civil Engineering Journal}, 8(7):1358--1368, 2022.

\bibitem{ballard2022contrastive}
Tristan Ballard and Gopal Erinjippurath.
\newblock Contrastive learning for climate model bias correction and super-resolution.
\newblock {\em arXiv preprint arXiv:2211.07555}, 2022.

\bibitem{bao2022physics}
Tianshu Bao, Shengyu Chen, Taylor~T Johnson, Peyman Givi, Shervin Sammak, and Xiaowei Jia.
\newblock Physics guided neural networks for spatio-temporal super-resolution of turbulent flows.
\newblock In {\em Uncertainty in Artificial Intelligence}, pages 118--128. PMLR, 2022.

\bibitem{bao2021partial}
Tianshu Bao, Xiaowei Jia, Jacob Zwart, Jeffrey Sadler, Alison Appling, Samantha Oliver, and Taylor~T Johnson.
\newblock Partial differential equation driven dynamic graph networks for predicting stream water temperature.
\newblock In {\em 2021 IEEE International Conference on Data Mining (ICDM)}, pages 11--20. IEEE, 2021.

\bibitem{bartholow2010stream}
John Bartholow.
\newblock Stream network and stream segment temperature models software.
\newblock Technical report, US Geological Survey, 2010.

\bibitem{bennett2021characterization}
Katrina~E Bennett, Satish Karra, and Velimir~V Vesselinov.
\newblock Characterization of extreme hydroclimate events in earth system models using ml/ai.
\newblock Technical report, Artificial Intelligence for Earth System Predictability (AI4ESP~…, 2021.

\bibitem{beven1992future}
Keith Beven and Andrew Binley.
\newblock The future of distributed models: model calibration and uncertainty prediction.
\newblock {\em Hydrological processes}, 6(3):279--298, 1992.

\bibitem{bocquet2023surrogate}
Marc Bocquet.
\newblock Surrogate modeling for the climate sciences dynamics with machine learning and data assimilation.
\newblock {\em Frontiers in Applied Mathematics and Statistics}, 9:1133226, 2023.

\bibitem{bolton2019applications}
Thomas Bolton and Laure Zanna.
\newblock Applications of deep learning to ocean data inference and subgrid parameterization.
\newblock {\em Journal of Advances in Modeling Earth Systems}, 11(1):376--399, 2019.

\bibitem{bommasani2021opportunities}
Rishi Bommasani, Drew~A Hudson, Ehsan Adeli, Russ Altman, Simran Arora, Sydney von Arx, Michael~S Bernstein, Jeannette Bohg, Antoine Bosselut, Emma Brunskill, et~al.
\newblock On the opportunities and risks of foundation models.
\newblock {\em arXiv preprint arXiv:2108.07258}, 2021.

\bibitem{boussif2022magnet}
Oussama Boussif, Yoshua Bengio, Loubna Benabbou, and Dan Assouline.
\newblock Magnet: Mesh agnostic neural pde solver.
\newblock {\em Advances in Neural Information Processing Systems}, 35:31972--31985, 2022.

\bibitem{brunton2017koopman}
Steven Brunton, Eurika Kaiser, and Nathan Kutz.
\newblock Koopman operator theory: Past, present, and future.
\newblock In {\em APS Division of Fluid Dynamics Meeting Abstracts}, pages L27--004, 2017.

\bibitem{caldwell2014statistical}
P.~Caldwell et~al.
\newblock Statistical significance of climate sensitivity predictors obtained by data mining.
\newblock {\em Geophysical Research Letters}, 41(5):1803--1808, 2014.

\bibitem{cang2018improving}
Ruijin Cang, Hechao Li, Hope Yao, Yang Jiao, and Yi~Ren.
\newblock Improving direct physical properties prediction of heterogeneous materials from imaging data via convolutional neural network and a morphology-aware generative model.
\newblock {\em Computational Materials Science}, 150:212--221, 2018.

\bibitem{chen2021heterogeneous}
Shengyu Chen, Alison Appling, Samantha Oliver, Hayley Corson-Dosch, Jordan Read, Jeffrey Sadler, Jacob Zwart, and Xiaowei Jia.
\newblock Heterogeneous stream-reservoir graph networks with data assimilation.
\newblock In {\em 2021 IEEE International Conference on Data Mining (ICDM)}, pages 1024--1029. IEEE, 2021.

\bibitem{chen2024reconstructing}
Shengyu Chen, Tianshu Bao, Peyman Givi, Can Zheng, and Xiaowei Jia.
\newblock Reconstructing turbulent flows using spatio-temporal physical dynamics.
\newblock {\em ACM Transactions on Intelligent Systems and Technology}, 15(1):1--18, 2024.

\bibitem{chen2024hossnet}
Shengyu Chen, Shihang Feng, Yao Huang, Zhou Lei, Xiaowei Jia, Youzuo Lin, and Estaben Rougier.
\newblock Hossnet: An efficient physics-guided neural network for simulating micro-crack propagation.
\newblock {\em Computational Materials Science}, 236:112846, 2024.

\bibitem{chen2023physics_stream}
Shengyu Chen, Nasrin Kalanat, Yiqun Xie, Sheng Li, Jacob~A Zwart, Jeffrey~M Sadler, Alison~P Appling, Samantha~K Oliver, Jordan~S Read, and Xiaowei Jia.
\newblock Physics-guided machine learning from simulated data with different physical parameters.
\newblock {\em Knowledge and Information Systems}, 65(8):3223--3250, 2023.

\bibitem{chen2021reconstructing}
Shengyu Chen, Shervin Sammak, Peyman Givi, Joseph~P Yurko, and Xiaowei Jia.
\newblock Reconstructing high-resolution turbulent flows using physics-guided neural networks.
\newblock In {\em 2021 IEEE International Conference on Big Data (Big Data)}, pages 1369--1379. IEEE, 2021.

\bibitem{chen2023physics}
Shengyu Chen, Yiqun Xie, Xiang Li, Xu~Liang, and Xiaowei Jia.
\newblock Physics-guided meta-learning method in baseflow prediction over large regions.
\newblock In {\em Proceedings of the 2023 SIAM International Conference on Data Mining (SDM)}, pages 217--225. SIAM, 2023.

\bibitem{chen2022physics}
Shengyu Chen, Jacob~A Zwart, and Xiaowei Jia.
\newblock Physics-guided graph meta learning for predicting water temperature and streamflow in stream networks.
\newblock In {\em Proceedings of the 28th ACM SIGKDD Conference on Knowledge Discovery and Data Mining}, pages 2752--2761, 2022.

\bibitem{chen2022videoinr}
Zeyuan Chen, Yinbo Chen, Jingwen Liu, Xingqian Xu, Vidit Goel, Zhangyang Wang, Humphrey Shi, and Xiaolong Wang.
\newblock Videoinr: Learning video implicit neural representation for continuous space-time super-resolution.
\newblock In {\em Proceedings of the IEEE/CVF Conference on Computer Vision and Pattern Recognition}, pages 2047--2057, 2022.

\bibitem{cohen2016group}
Taco Cohen and Max Welling.
\newblock Group equivariant convolutional networks.
\newblock In {\em International conference on machine learning}, pages 2990--2999. PMLR, 2016.

\bibitem{cohen2016steerable}
Taco~S Cohen and Max Welling.
\newblock Steerable cnns.
\newblock {\em arXiv preprint arXiv:1612.08498}, 2016.

\bibitem{croitoru2023diffusion}
Florinel-Alin Croitoru, Vlad Hondru, Radu~Tudor Ionescu, and Mubarak Shah.
\newblock Diffusion models in vision: A survey.
\newblock {\em IEEE Transactions on Pattern Analysis and Machine Intelligence}, 2023.

\bibitem{daw2023mitigating}
Arka Daw, Jie Bu, Sifan Wang, Paris Perdikaris, and Anuj Karpatne.
\newblock Mitigating propagation failures in physics-informed neural networks using retain-resample-release (r3) sampling.
\newblock In {\em Proceedings of the 40th International Conference on Machine Learning}, ICML'23. JMLR.org, 2023.

\bibitem{daw2022physics}
Arka Daw, Anuj Karpatne, William~D Watkins, Jordan~S Read, and Vipin Kumar.
\newblock Physics-guided neural networks (pgnn): An application in lake temperature modeling.
\newblock In {\em Knowledge Guided Machine Learning}, pages 353--372. Chapman and Hall/CRC, 2022.

\bibitem{daw2020physics}
Arka Daw, R~Quinn Thomas, Cayelan~C Carey, Jordan~S Read, Alison~P Appling, and Anuj Karpatne.
\newblock Physics-guided architecture (pga) of neural networks for quantifying uncertainty in lake temperature modeling.
\newblock In {\em Proceedings of the 2020 SIAM International Conference on Data Mining}, pages 532--540. SIAM, 2020.

\bibitem{imagenet_cvpr09}
J.~Deng, W.~Dong, R.~Socher, L.-J. Li, K.~Li, and L.~Fei-Fei.
\newblock {ImageNet: A Large-Scale Hierarchical Image Database}.
\newblock In {\em CVPR09}, 2009.

\bibitem{dugdale2017river}
Stephen~J Dugdale, David~M Hannah, and Iain~A Malcolm.
\newblock River temperature modelling: A review of process-based approaches and future directions.
\newblock {\em Earth-Science Reviews}, 175:97--113, 2017.

\bibitem{duraisamy2019turbulence}
Karthik Duraisamy, Gianluca Iaccarino, and Heng Xiao.
\newblock Turbulence modeling in the age of data.
\newblock {\em Annual review of fluid mechanics}, 51:357--377, 2019.

\bibitem{elhamod2022cophy}
Mohannad Elhamod, Jie Bu, Christopher Singh, Matthew Redell, Abantika Ghosh, Viktor Podolskiy, Wei-Cheng Lee, and Anuj Karpatne.
\newblock Cophy-pgnn: Learning physics-guided neural networks with competing loss functions for solving eigenvalue problems.
\newblock {\em ACM Transactions on Intelligent Systems and Technology}, 13(6):1--23, 2022.

\bibitem{engl1996regularization}
Heinz~Werner Engl, Martin Hanke, and Andreas Neubauer.
\newblock {\em Regularization of inverse problems}, volume 375.
\newblock Springer Science \& Business Media, 1996.

\bibitem{faghmous2014theory}
J.~H. Faghmous, A.~Banerjee, S.~Shekhar, M.~Steinbach, V.~Kumar, A.~R. Ganguly, and N.~Samatova.
\newblock Theory-guided data science for climate change.
\newblock {\em Computer}, 47(11):74--78, 2014.

\bibitem{fan2020optimal}
Yue Fan, Wenxi Lu, Tiansheng Miao, Yongkai An, Jiuhui Li, and Jiannan Luo.
\newblock Optimal design of groundwater pollution monitoring network based on the svr surrogate model under uncertainty.
\newblock {\em Environmental Science and Pollution Research}, 27:24090--24102, 2020.

\bibitem{fatichi2016overview}
Simone Fatichi, Enrique~R Vivoni, Fred~L Ogden, Valeriy~Y Ivanov, Benjamin Mirus, David Gochis, Charles~W Downer, Matteo Camporese, Jason~H Davison, Brian Ebel, et~al.
\newblock An overview of current applications, challenges, and future trends in distributed process-based models in hydrology.
\newblock {\em Journal of Hydrology}, 537:45--60, 2016.

\bibitem{feng2022differentiable}
Dapeng Feng, Jiangtao Liu, Kathryn Lawson, and Chaopeng Shen.
\newblock Differentiable, learnable, regionalized process-based models with multiphysical outputs can approach state-of-the-art hydrologic prediction accuracy.
\newblock {\em Water Resources Research}, 58(10):e2022WR032404, 2022.

\bibitem{jacob2007first}
J~Fish and T~Belytschko.
\newblock {\em A first course in finite elements}.
\newblock Wiley, 2007.

\bibitem{fisher2023opening}
Susannah Fisher.
\newblock Opening up new geographical ontologies around adapting to climate change.
\newblock {\em Tijdschrift voor economische en sociale geografie}, 114(2):79--85, 2023.

\bibitem{furtney2022surrogate}
JK~Furtney, C~Thielsen, W~Fu, and Romain Le~Goc.
\newblock Surrogate models in rock and soil mechanics: Integrating numerical modeling and machine learning.
\newblock {\em Rock Mechanics and Rock Engineering}, pages 1--15, 2022.

\bibitem{gao2023prediff}
Zhihan Gao, Xingjian Shi, Boran Han, Hao Wang, Xiaoyong Jin, Danielle Maddix, Yi~Zhu, Mu~Li, and Yuyang Wang.
\newblock Prediff: Precipitation nowcasting with latent diffusion models.
\newblock {\em arXiv preprint arXiv:2307.10422}, 2023.

\bibitem{ghorbanidehno2020recent}
Hojat Ghorbanidehno, Amalia Kokkinaki, Jonghyun Lee, and Eric Darve.
\newblock Recent developments in fast and scalable inverse modeling and data assimilation methods in hydrology.
\newblock {\em Journal of Hydrology}, 591:125266, 2020.

\bibitem{ghosh2022physics}
Abantika Ghosh, Mohannad Elhamod, Jie Bu, Wei-Cheng Lee, Anuj Karpatne, and Viktor~A Podolskiy.
\newblock Physics-informed machine learning for optical modes in composites.
\newblock {\em Advanced Photonics Research}, 3(11):2200073, 2022.

\bibitem{ghosh2022robust}
Rahul Ghosh, Arvind Renganathan, Kshitij Tayal, Xiang Li, Ankush Khandelwal, Xiaowei Jia, Christopher Duffy, John Nieber, and Vipin Kumar.
\newblock Robust inverse framework using knowledge-guided self-supervised learning: An application to hydrology.
\newblock In {\em Proceedings of the 28th ACM SIGKDD Conference on Knowledge Discovery and Data Mining}, pages 465--474, 2022.

\bibitem{ghosh2023entity}
Rahul Ghosh, Haoyu Yang, Ankush Khandelwal, Erhu He, Arvind Renganathan, Somya Sharma, Xiaowei Jia, and Vipin Kumar.
\newblock Entity aware modelling: A survey.
\newblock {\em arXiv preprint arXiv:2302.08406}, 2023.

\bibitem{goodfellow2014generative}
Ian Goodfellow, Jean Pouget-Abadie, Mehdi Mirza, Bing Xu, David Warde-Farley, Sherjil Ozair, Aaron Courville, and Yoshua Bengio.
\newblock Generative adversarial nets.
\newblock In {\em NIPS}, pages 2672--2680, 2014.

\bibitem{goswami2022physics}
Somdatta Goswami, Aniruddha Bora, Yue Yu, and George~Em Karniadakis.
\newblock Physics-informed neural operators.
\newblock {\em arXiv preprint arXiv:2207.05748}, 2022.

\bibitem{guevara2019downscaling}
Mario Guevara and Rodrigo Vargas.
\newblock Downscaling satellite soil moisture using geomorphometry and machine learning.
\newblock {\em PloS One}, 14(9):e0219639, 2019.

\bibitem{guo2022knowledge}
Yinan Guo, Guoyu Chen, Min Jiang, Dunwei Gong, and Jing Liang.
\newblock A knowledge guided transfer strategy for evolutionary dynamic multiobjective optimization.
\newblock {\em IEEE Transactions on Evolutionary Computation}, 2022.

\bibitem{han2023deeporyza}
Jingye Han, Liangsheng Shi, Christos Pylianidis, Qi~Yang, and Ioannis~N Athanasiadis.
\newblock Deeporyza: A knowledge guided machine learning model for rice growth simulation.
\newblock In {\em 2nd AAAI Workshop on AI for Agriculture and Food Systems}, 2023.

\bibitem{hanson2020predicting}
Paul~C Hanson, Aviah~B Stillman, Xiaowei Jia, Anuj Karpatne, Hilary~A Dugan, Cayelan~C Carey, Joseph Stachelek, Nicole~K Ward, Yu~Zhang, Jordan~S Read, et~al.
\newblock Predicting lake surface water phosphorus dynamics using process-guided machine learning.
\newblock {\em Ecological Modelling}, 430:109136, 2020.

\bibitem{harilal2021augmented}
Nidhin Harilal, Mayank Singh, and Udit Bhatia.
\newblock Augmented convolutional lstms for generation of high-resolution climate change projections.
\newblock {\em IEEE Access}, 9:25208--25218, 2021.

\bibitem{he2023physics}
Erhu He, Yiqun Xie, Licheng Liu, Weiye Chen, Zhenong Jin, and Xiaowei Jia.
\newblock Physics guided neural networks for time-aware fairness: an application in crop yield prediction.
\newblock In {\em Proceedings of the AAAI Conference on Artificial Intelligence}, volume~37, pages 14223--14231, 2023.

\bibitem{galactica}
Will~Douglas Heaven.
\newblock Why {M}eta’s latest large language model survived only three days online.
\newblock {\em MIT Technology Review}, 2022.

\bibitem{hettige2024airphynet}
Kethmi~Hirushini Hettige, Jiahao Ji, Shili Xiang, Cheng Long, Gao Cong, and Jingyuan Wang.
\newblock Airphynet: Harnessing physics-guided neural networks for air quality prediction.
\newblock {\em arXiv preprint arXiv:2402.03784}, 2024.

\bibitem{hill2006effective}
Mary~C Hill and Claire~R Tiedeman.
\newblock {\em Effective groundwater model calibration: with analysis of data, sensitivities, predictions, and uncertainty}.
\newblock John Wiley \& Sons, 2006.

\bibitem{himes2022accurate}
Michael~D Himes, Joseph Harrington, Adam~D Cobb, At{\i}l{\i}m~G{\"u}ne{\c{s}} Baydin, Frank Soboczenski, Molly~D O’Beirne, Simone Zorzan, David~C Wright, Zacchaeus Scheffer, Shawn~D Domagal-Goldman, et~al.
\newblock Accurate machine-learning atmospheric retrieval via a neural-network surrogate model for radiative transfer.
\newblock {\em The Planetary Science Journal}, 3(4):91, 2022.

\bibitem{hipsey2019general}
Matthew~R Hipsey, Louise~C Bruce, Casper Boon, Brendan Busch, Cayelan~C Carey, David~P Hamilton, Paul~C Hanson, Jordan~S Read, Eduardo de~Sousa, Michael Weber, et~al.
\newblock A general lake model (glm 3.0) for linking with high-frequency sensor data from the global lake ecological observatory network (gleon).
\newblock {\em Geoscientific Model Development}, 12(1):473--523, 2019.

\bibitem{hipsey2014glm}
MR~Hipsey, LC~Bruce, and DP~Hamilton.
\newblock Glm—general lake model: Model overview and user information.
\newblock {\em Perth (Australia): University of Western Australia Technical Manual}, 2014.

\bibitem{ho2020denoising}
Jonathan Ho, Ajay Jain, and Pieter Abbeel.
\newblock Denoising diffusion probabilistic models.
\newblock {\em Advances in neural information processing systems}, 33:6840--6851, 2020.

\bibitem{iqbal2022survey}
Touseef Iqbal and Shaima Qureshi.
\newblock The survey: Text generation models in deep learning.
\newblock {\em Journal of King Saud University-Computer and Information Sciences}, 34(6):2515--2528, 2022.

\bibitem{ishitsuka2023physics}
Kazuya Ishitsuka and Weiren Lin.
\newblock Physics-informed neural network for inverse modeling of natural-state geothermal systems.
\newblock {\em Applied Energy}, 337:120855, 2023.

\bibitem{jebeile2021understanding}
Julie Jebeile, Vincent Lam, and Tim R{\"a}z.
\newblock Understanding climate change with statistical downscaling and machine learning.
\newblock {\em Synthese}, 199:1877--1897, 2021.

\bibitem{jia2022modeling}
Xiaowei Jia, Shengyu Chen, Yiqun Xie, Haoyu Yang, Alison Appling, Samantha Oliver, and Zhe Jiang.
\newblock Modeling reservoir release using pseudo-prospective learning and physical simulations to predict water temperature.
\newblock In {\em Proceedings of the 2022 SIAM International Conference on Data Mining (SDM)}, pages 91--99. SIAM, 2022.

\bibitem{jia2023physics}
Xiaowei Jia, Shengyu Chen, Can Zheng, Yiqun Xie, Zhe Jiang, and Nasrin Kalanat.
\newblock Physics-guided graph diffusion network for combining heterogeneous simulated data: An application in predicting stream water temperature.
\newblock In {\em Proceedings of the 2023 SIAM International Conference on Data Mining (SDM)}, pages 361--369. SIAM, 2023.

\bibitem{jia2021simlr}
Xiaowei Jia et~al.
\newblock Physics-guided machine learning from simulation data: an application in modeling lake and river systems.
\newblock In {\em ICDM}, 2021.

\bibitem{jia2019recurrent}
Xiaowei Jia, Mengdie Wang, Ankush Khandelwal, Anuj Karpatne, and Vipin Kumar.
\newblock Recurrent generative networks for multi-resolution satellite data: An application in cropland monitoring.
\newblock In {\em IJCAI}, 2019.

\bibitem{jia2019physics}
Xiaowei Jia, Jared Willard, Anuj Karpatne, Jordan Read, Jacob Zwart, Michael Steinbach, and Vipin Kumar.
\newblock Physics guided rnns for modeling dynamical systems: A case study in simulating lake temperature profiles.
\newblock In {\em Proceedings of the 2019 SIAM International Conference on Data Mining}, pages 558--566. SIAM, 2019.

\bibitem{jia2021physics}
Xiaowei Jia, Jared Willard, Anuj Karpatne, Jordan~S Read, Jacob~A Zwart, Michael Steinbach, and Vipin Kumar.
\newblock Physics-guided machine learning for scientific discovery: An application in simulating lake temperature profiles.
\newblock {\em ACM/IMS Transactions on Data Science}, 2(3):1--26, 2021.

\bibitem{jia2021physics_pgrgrn}
Xiaowei Jia, Jacob Zwart, Jeffrey Sadler, Alison Appling, Samantha Oliver, Steven Markstrom, Jared Willard, Shaoming Xu, Michael Steinbach, Jordan Read, et~al.
\newblock Physics-guided recurrent graph model for predicting flow and temperature in river networks.
\newblock In {\em Proceedings of the 2021 SIAM International Conference on Data Mining (SDM)}, pages 612--620. SIAM, 2021.

\bibitem{jiang2023efficient}
Peishi Jiang, Zhao Yang, Jiali Wang, Chenfu Huang, Pengfei Xue, TC~Chakraborty, Xingyuan Chen, and Yun Qian.
\newblock Efficient super-resolution of near-surface climate modeling using the fourier neural operator.
\newblock {\em Journal of Advances in Modeling Earth Systems}, 15(7):e2023MS003800, 2023.

\bibitem{jin2021unsupervised}
Peng Jin, Xitong Zhang, Yinpeng Chen, Sharon~Xiaolei Huang, Zicheng Liu, and Youzuo Lin.
\newblock Unsupervised learning of full-waveform inversion: Connecting cnn and partial differential equation in a loop.
\newblock {\em arXiv preprint arXiv:2110.07584}, 2021.

\bibitem{karniadakis2021physics}
George~Em Karniadakis, Ioannis~G Kevrekidis, Lu~Lu, Paris Perdikaris, Sifan Wang, and Liu Yang.
\newblock Physics-informed machine learning.
\newblock {\em Nature Reviews Physics}, 3(6):422--440, 2021.

\bibitem{tgds}
Anuj Karpatne, Gowtham Atluri, James~H Faghmous, Michael Steinbach, Arindam Banerjee, Auroop Ganguly, Shashi Shekhar, Nagiza Samatova, and Vipin Kumar.
\newblock Theory-guided data science: A new paradigm for scientific discovery from data.
\newblock {\em IEEE Transactions on Knowledge and Data Engineering}, 29(10):2318--2331, 2017.

\bibitem{karpatne2022knowledge}
Anuj Karpatne, Ramakrishnan Kannan, and Vipin Kumar.
\newblock {\em Knowledge Guided Machine Learning: Accelerating Discovery Using Scientific Knowledge and Data}.
\newblock CRC Press, 2022.

\bibitem{karpatne2017physics}
Anuj Karpatne, William Watkins, Jordan Read, and Vipin Kumar.
\newblock Physics-guided neural networks (pgnn): An application in lake temperature modeling.
\newblock {\em arXiv preprint arXiv:1710.11431}, 2017.

\bibitem{kashinath2021physics}
K~Kashinath, M~Mustafa, A~Albert, JL~Wu, C~Jiang, S~Esmaeilzadeh, K~Azizzadenesheli, R~Wang, A~Chattopadhyay, A~Singh, et~al.
\newblock Physics-informed machine learning: case studies for weather and climate modelling.
\newblock {\em Philosophical Transactions of the Royal Society A}, 379(2194):20200093, 2021.

\bibitem{kingma2013auto}
Diederik~P Kingma and Max Welling.
\newblock Auto-encoding variational bayes.
\newblock {\em arXiv preprint arXiv:1312.6114}, 2013.

\bibitem{kobyzev2020normalizing}
Ivan Kobyzev, Simon~JD Prince, and Marcus~A Brubaker.
\newblock Normalizing flows: An introduction and review of current methods.
\newblock {\em IEEE transactions on pattern analysis and machine intelligence}, 43(11):3964--3979, 2020.

\bibitem{kovachki2021neural}
Nikola Kovachki, Zongyi Li, Burigede Liu, Kamyar Azizzadenesheli, Kaushik Bhattacharya, Andrew Stuart, and Anima Anandkumar.
\newblock Neural operator: Learning maps between function spaces.
\newblock {\em arXiv preprint arXiv:2108.08481}, 2021.

\bibitem{kraft2022towards}
Basil Kraft, Martin Jung, Marco K{\"o}rner, Sujan Koirala, and Markus Reichstein.
\newblock Towards hybrid modeling of the global hydrological cycle.
\newblock {\em Hydrology and Earth System Sciences}, 26(6):1579--1614, 2022.

\bibitem{kratzert2019towards}
Frederik Kratzert, Daniel Klotz, Guy Shalev, G{\"u}nter Klambauer, Sepp Hochreiter, and Grey Nearing.
\newblock Towards learning universal, regional, and local hydrological behaviors via machine learning applied to large-sample datasets.
\newblock {\em Hydrology and Earth System Sciences}, 23(12):5089--5110, 2019.

\bibitem{krenn2022scientific}
Mario Krenn, Robert Pollice, Si~Yue Guo, Matteo Aldeghi, Alba Cervera-Lierta, Pascal Friederich, Gabriel dos Passos~Gomes, Florian H{\"a}se, Adrian Jinich, AkshatKumar Nigam, et~al.
\newblock On scientific understanding with artificial intelligence.
\newblock {\em Nature Reviews Physics}, 4(12):761--769, 2022.

\bibitem{kumar2022fine}
Ananya Kumar, Aditi Raghunathan, Robbie Jones, Tengyu Ma, and Percy Liang.
\newblock Fine-tuning can distort pretrained features and underperform out-of-distribution.
\newblock {\em arXiv preprint arXiv:2202.10054}, 2022.

\bibitem{kumar2023advanced}
Vijendra Kumar, Naresh Kedam, Kul~Vaibhav Sharma, Darshan~J Mehta, and Tommaso Caloiero.
\newblock Advanced machine learning techniques to improve hydrological prediction: A comparative analysis of streamflow prediction models.
\newblock {\em Water}, 15(14):2572, 2023.

\bibitem{lacoste2023geo}
Alexandre Lacoste, Nils Lehmann, Pau Rodriguez, Evan~David Sherwin, Hannah Kerner, Bj{\"o}rn L{\"u}tjens, Jeremy~Andrew Irvin, David Dao, Hamed Alemohammad, Alexandre Drouin, et~al.
\newblock Geo-bench: Toward foundation models for earth monitoring.
\newblock {\em arXiv preprint arXiv:2306.03831}, 2023.

\bibitem{lambhate2020super}
Devyani Lambhate and Deepak~N Subramani.
\newblock Super-resolution of sea surface temperature satellite images.
\newblock In {\em Global Oceans 2020: Singapore--US Gulf Coast}, pages 1--7. IEEE, 2020.

\bibitem{larsson2017colorization}
Gustav Larsson, Michael Maire, and Gregory Shakhnarovich.
\newblock Colorization as a proxy task for visual understanding.
\newblock In {\em Proceedings of the IEEE Conference on Computer Vision and Pattern Recognition}, pages 6874--6883, 2017.

\bibitem{Lazer2014}
David Lazer, Ryan Kennedy, Gary King, and Alessandro Vespignani.
\newblock {The Parable of Google Flu: Traps in Big Data Analysis}.
\newblock {\em Science (New York, N.Y.)}, 343(6176):1203--5, March 2014.

\bibitem{le2020contrastive}
Phuc~H Le-Khac, Graham Healy, and Alan~F Smeaton.
\newblock Contrastive representation learning: A framework and review.
\newblock {\em Ieee Access}, 8:193907--193934, 2020.

\bibitem{li2022regionalization}
Xiang Li, Ankush Khandelwal, Xiaowei Jia, Kelly Cutler, Rahul Ghosh, Arvind Renganathan, Shaoming Xu, Kshitij Tayal, John Nieber, Christopher Duffy, et~al.
\newblock Regionalization in a global hydrologic deep learning model: from physical descriptors to random vectors.
\newblock {\em Water Resources Research}, 58(8):e2021WR031794, 2022.

\bibitem{li2022fourier}
Zongyi Li, Daniel~Zhengyu Huang, Burigede Liu, and Anima Anandkumar.
\newblock Fourier neural operator with learned deformations for pdes on general geometries.
\newblock {\em arXiv preprint arXiv:2207.05209}, 2022.

\bibitem{li2020fourier}
Zongyi Li, Nikola Kovachki, Kamyar Azizzadenesheli, Burigede Liu, Kaushik Bhattacharya, Andrew Stuart, and Anima Anandkumar.
\newblock Fourier neural operator for parametric partial differential equations.
\newblock {\em arXiv preprint arXiv:2010.08895}, 2020.

\bibitem{li2021physics}
Zongyi Li, Hongkai Zheng, Nikola Kovachki, David Jin, Haoxuan Chen, Burigede Liu, Kamyar Azizzadenesheli, and Anima Anandkumar.
\newblock Physics-informed neural operator for learning partial differential equations.
\newblock {\em arXiv preprint arXiv:2111.03794}, 2021.

\bibitem{licheng2023knowledge}
LIU Licheng, Wang Zhou, Kaiyu Guan, Bin Peng, Chongya Jiang, Jinyun Tang, Sheng Wang, Robert Grant, Symon Mezbahuddin, Xiaowei Jia, et~al.
\newblock Knowledge-based artificial intelligence for agroecosystem carbon budget and crop yield estimation.
\newblock {\em Authorea Preprints}, 2023.

\bibitem{licheng2022estimating}
LIU Licheng, Wang Zhou, Zhenong Jin, Jinyun Tang, Xiaowei Jia, Chongya Jiang, Kaiyu Guan, Bin Peng, Shaoming Xu, Yufeng Yang, et~al.
\newblock Estimating the autotrophic and heterotrophic respiration in the us crop fields using knowledge guided machine learning.
\newblock {\em Authorea Preprints}, 2022.

\bibitem{lienen2022learning}
Marten Lienen and Stephan G{\"u}nnemann.
\newblock Learning the dynamics of physical systems from sparse observations with finite element networks.
\newblock {\em arXiv preprint arXiv:2203.08852}, 2022.

\bibitem{lippe2023pde}
Phillip Lippe, Bastiaan~S Veeling, Paris Perdikaris, Richard~E Turner, and Johannes Brandstetter.
\newblock Pde-refiner: Achieving accurate long rollouts with neural pde solvers.
\newblock {\em arXiv preprint arXiv:2308.05732}, 2023.

\bibitem{liu2021study}
Lei Liu and Xue-yi You.
\newblock Study of water quality response to water transfer patterns in the receiving basin and surrogate model.
\newblock {\em Environmental Science and Pollution Research}, pages 1--19, 2021.

\bibitem{liu2022kgml}
Licheng Liu, Shaoming Xu, Jinyun Tang, Kaiyu Guan, Timothy~J Griffis, Matthew~D Erickson, Alexander~L Frie, Xiaowei Jia, Taegon Kim, Lee~T Miller, et~al.
\newblock Kgml-ag: a modeling framework of knowledge-guided machine learning to simulate agroecosystems: a case study of estimating n 2 o emission using data from mesocosm experiments.
\newblock {\em Geoscientific Model Development}, 15(7):2839--2858, 2022.

\bibitem{liu2024knowledge}
Licheng Liu, Wang Zhou, Kaiyu Guan, Bin Peng, Shaoming Xu, Jinyun Tang, Qing Zhu, Jessica Till, Xiaowei Jia, Chongya Jiang, et~al.
\newblock Knowledge-guided machine learning can improve carbon cycle quantification in agroecosystems.
\newblock {\em Nature Communications}, 15(1):357, 2024.

\bibitem{liu2020downscaling}
Yangxiaoyue Liu, Xiaolin Xia, Ling Yao, Wenlong Jing, Chenghu Zhou, Wumeng Huang, Yong Li, and Ji~Yang.
\newblock Downscaling satellite retrieved soil moisture using regression tree-based machine learning algorithms over southwest france.
\newblock {\em Earth and Space Science}, 7(10):e2020EA001267, 2020.

\bibitem{ljung1998system}
Lennart Ljung.
\newblock System identification.
\newblock In {\em Signal analysis and prediction}, pages 163--173. Springer, 1998.

\bibitem{lorsung2024picl}
Cooper Lorsung and Amir~Barati Farimani.
\newblock Picl: Physics informed contrastive learning for partial differential equations.
\newblock {\em arXiv preprint arXiv:2401.16327}, 2024.

\bibitem{lu2019deeponet}
Lu~Lu, Pengzhan Jin, and George~Em Karniadakis.
\newblock Deeponet: Learning nonlinear operators for identifying differential equations based on the universal approximation theorem of operators.
\newblock {\em arXiv preprint arXiv:1910.03193}, 2019.

\bibitem{lu2021learning}
Lu~Lu, Pengzhan Jin, Guofei Pang, Zhongqiang Zhang, and George~Em Karniadakis.
\newblock Learning nonlinear operators via deeponet based on the universal approximation theorem of operators.
\newblock {\em Nature machine intelligence}, 3(3):218--229, 2021.

\bibitem{luo2023free}
Shiyuan Luo, Juntong Ni, Shengyu Chen, Runlong Yu, Yiqun Xie, Licheng Liu, Zhenong Jin, Huaxiu Yao, and Xiaowei Jia.
\newblock Free: The foundational semantic recognition for modeling environmental ecosystems.
\newblock {\em arXiv preprint arXiv:2311.10255}, 2023.

\bibitem{luo2023physics}
Yingtao Luo, Qiang Liu, Yuntian Chen, Wenbo Hu, Tian Tian, and Jun Zhu.
\newblock Physics-guided discovery of highly nonlinear parametric partial differential equations.
\newblock In {\em Proceedings of the 29th ACM SIGKDD Conference on Knowledge Discovery and Data Mining}, pages 1595--1607, 2023.

\bibitem{lutjens2022multiscale}
Bj{\"o}rn L{\"u}tjens, Catherine~H Crawford, Campbell~D Watson, Christopher Hill, and Dava Newman.
\newblock Multiscale neural operator: Learning fast and grid-independent pde solvers.
\newblock {\em arXiv preprint arXiv:2207.11417}, 2022.

\bibitem{lynch2008origins}
Peter Lynch.
\newblock The origins of computer weather prediction and climate modeling.
\newblock {\em Journal of computational physics}, 227(7):3431--3444, 2008.

\bibitem{mai2023opportunities}
Gengchen Mai, Weiming Huang, Jin Sun, Suhang Song, Deepak Mishra, Ninghao Liu, Song Gao, Tianming Liu, Gao Cong, Yingjie Hu, et~al.
\newblock On the opportunities and challenges of foundation models for geospatial artificial intelligence.
\newblock {\em arXiv preprint arXiv:2304.06798}, 2023.

\bibitem{makela2000process}
Annikki M{\"a}kel{\"a}, Joe Landsberg, Alan~R Ek, Thomas~E Burk, Michael Ter-Mikaelian, G{\"o}ran~I {\AA}gren, Chadwick~D Oliver, and Pasi Puttonen.
\newblock Process-based models for forest ecosystem management: current state of the art and challenges for practical implementation.
\newblock {\em Tree physiology}, 20(5-6):289--298, 2000.

\bibitem{marcus2014eight}
Gary Marcus and Ernest Davis.
\newblock Eight (no, nine!) problems with big data.
\newblock {\em The New York Times}, 6(04):2014, 2014.

\bibitem{mccabe2023multiple}
Michael McCabe, Bruno R{\'e}galdo-Saint Blancard, Liam~Holden Parker, Ruben Ohana, Miles Cranmer, Alberto Bietti, Michael Eickenberg, Siavash Golkar, Geraud Krawezik, Francois Lanusse, et~al.
\newblock Multiple physics pretraining for physical surrogate models.
\newblock {\em arXiv preprint arXiv:2310.02994}, 2023.

\bibitem{muralidhar2018incorporating}
Nikhil Muralidhar, Mohammad~Raihanul Islam, Manish Marwah, Anuj Karpatne, and Naren Ramakrishnan.
\newblock Incorporating prior domain knowledge into deep neural networks.
\newblock In {\em 2018 IEEE International Conference on Big Data (Big Data)}, pages 36--45. IEEE, 2018.

\bibitem{nguyen2023climax}
Tung Nguyen, Johannes Brandstetter, Ashish Kapoor, Jayesh~K Gupta, and Aditya Grover.
\newblock Climax: A foundation model for weather and climate.
\newblock {\em arXiv preprint arXiv:2301.10343}, 2023.

\bibitem{norvig2009natural}
Peter Norvig.
\newblock Natural language corpus data.
\newblock {\em Beautiful data}, pages 219--242, 2009.

\bibitem{openai2023gpt}
R~OpenAI.
\newblock Gpt-4 technical report. arxiv 2303.08774.
\newblock {\em View in Article}, 2, 2023.

\bibitem{park2022downscaling}
Sungwon Park, Karandeep Singh, Arjun Nellikkattil, Elke Zeller, Tung~Duong Mai, and Meeyoung Cha.
\newblock Downscaling earth system models with deep learning.
\newblock In {\em Proceedings of the 28th ACM SIGKDD Conference on Knowledge Discovery and Data Mining}, pages 3733--3742, 2022.

\bibitem{pathak2016context}
Deepak Pathak, Philipp Krahenbuhl, Jeff Donahue, Trevor Darrell, and Alexei~A Efros.
\newblock Context encoders: Feature learning by inpainting.
\newblock In {\em Proceedings of the IEEE conference on computer vision and pattern recognition}, pages 2536--2544, 2016.

\bibitem{pearl2019seven}
Judea Pearl.
\newblock The seven tools of causal inference, with reflections on machine learning.
\newblock {\em Communications of the ACM}, 62(3):54--60, 2019.

\bibitem{pearl2018book}
Judea Pearl and Dana Mackenzie.
\newblock {\em The book of why: the new science of cause and effect}.
\newblock Basic books, 2018.

\bibitem{pilozzi2018machine}
Laura Pilozzi, Francis~A Farrelly, Giulia Marcucci, and Claudio Conti.
\newblock Machine learning inverse problem for topological photonics.
\newblock {\em Communications Physics}, 1(1):1--7, 2018.

\bibitem{raissi2018deep}
Maziar Raissi.
\newblock Deep hidden physics models: Deep learning of nonlinear partial differential equations.
\newblock {\em The Journal of Machine Learning Research}, 19(1):932--955, 2018.

\bibitem{raissi2019physics}
Maziar Raissi, Paris Perdikaris, and George~E Karniadakis.
\newblock Physics-informed neural networks: A deep learning framework for solving forward and inverse problems involving nonlinear partial differential equations.
\newblock {\em Journal of Computational Physics}, 378:686--707, 2019.

\bibitem{raissi2017physics1}
Maziar Raissi, Paris Perdikaris, and George~Em Karniadakis.
\newblock Physics informed deep learning (part i): Data-driven solutions of nonlinear partial differential equations.
\newblock {\em arXiv preprint arXiv:1711.10561}, 2017.

\bibitem{raissi2017physics2}
Maziar Raissi, Paris Perdikaris, and George~Em Karniadakis.
\newblock Physics informed deep learning (part ii): Data-driven discovery of nonlinear partial differential equations.
\newblock {\em arXiv preprint arXiv:1711.10566}, 2017.

\bibitem{rasp2018deep}
Stephan Rasp, Michael~S Pritchard, and Pierre Gentine.
\newblock Deep learning to represent subgrid processes in climate models.
\newblock {\em Proceedings of the National Academy of Sciences}, 115(39):9684--9689, 2018.

\bibitem{razavi2012review}
Saman Razavi, Bryan~A Tolson, and Donald~H Burn.
\newblock Review of surrogate modeling in water resources.
\newblock {\em Water Resources Research}, 48(7), 2012.

\bibitem{read2019process}
Jordan~S Read, Xiaowei Jia, Jared Willard, Alison~P Appling, Jacob~A Zwart, Samantha~K Oliver, Anuj Karpatne, Gretchen~JA Hansen, Paul~C Hanson, William Watkins, et~al.
\newblock Process-guided deep learning predictions of lake water temperature.
\newblock {\em Water Resources Research}, 55(11):9173--9190, 2019.

\bibitem{reichstein2019deep}
Markus Reichstein, Gustau Camps-Valls, Bjorn Stevens, Martin Jung, Joachim Denzler, Nuno Carvalhais, et~al.
\newblock Deep learning and process understanding for data-driven earth system science.
\newblock {\em Nature}, 566(7743):195--204, 2019.

\bibitem{rudy2017data}
Samuel~H Rudy, Steven~L Brunton, Joshua~L Proctor, and J~Nathan Kutz.
\newblock Data-driven discovery of partial differential equations.
\newblock {\em Science advances}, 3(4):e1602614, 2017.

\bibitem{sabzehee2023enhancing}
F~Sabzehee, AR~Amiri-Simkooei, S~Iran-Pour, BD~Vishwakarma, and R~Kerachian.
\newblock Enhancing spatial resolution of grace-derived groundwater storage anomalies in urmia catchment using machine learning downscaling methods.
\newblock {\em Journal of Environmental Management}, 330:117180, 2023.

\bibitem{satorras2021n}
V{\i}ctor~Garcia Satorras, Emiel Hoogeboom, and Max Welling.
\newblock E (n) equivariant graph neural networks.
\newblock In {\em International conference on machine learning}, pages 9323--9332. PMLR, 2021.

\bibitem{schneider2022machine}
Raphael Schneider, Julian Koch, Lars Troldborg, Hans~J{\o}rgen Henriksen, and Simon Stisen.
\newblock Machine-learning-based downscaling of modelled climate change impacts on groundwater table depth.
\newblock {\em Hydrology and Earth System Sciences}, 26(22):5859--5877, 2022.

\bibitem{sharma2023knowledge}
Somya Sharma, Swati Sharma, Licheng Liu, Rishabh Tushir, Andy Neal, Robert Ness, John Crawford, Emre Kiciman, and Ranveer Chandra.
\newblock Knowledge guided representation learning and causal structure learning in soil science.
\newblock {\em arXiv preprint arXiv:2306.09302}, 2023.

\bibitem{shen2023differentiable}
Chaopeng Shen, Alison~P Appling, Pierre Gentine, Toshiyuki Bandai, Hoshin Gupta, Alexandre Tartakovsky, Marco Baity-Jesi, Fabrizio Fenicia, Daniel Kifer, Li~Li, et~al.
\newblock Differentiable modelling to unify machine learning and physical models for geosciences.
\newblock {\em Nature Reviews Earth \& Environment}, 4(8):552--567, 2023.

\bibitem{sit2023spatial}
Muhammed Sit, Bekir~Zahit Demiray, and Ibrahim Demir.
\newblock Spatial downscaling of streamflow data with attention based spatio-temporal graph convolutional networks.
\newblock 2023.

\bibitem{tarantola2005inverse}
Albert Tarantola.
\newblock {\em Inverse problem theory and methods for model parameter estimation}.
\newblock SIAM, 2005.

\bibitem{tayal2022invertibility}
Kshitij Tayal, Xiaowei Jia, Rahul Ghosh, Jared Willard, Jordan Read, and Vipin Kumar.
\newblock Invertibility aware integration of static and time-series data: An application to lake temperature modeling.
\newblock In {\em Proceedings of the 2022 SIAM International Conference on Data Mining (SDM)}, pages 702--710. SIAM, 2022.

\bibitem{tayal2023koopman}
Kshitij Tayal, Arvind Renganathan, Rahul Ghosh, Xiaowei Jia, and Vipin Kumar.
\newblock Koopman invertible autoencoder: Leveraging forward and backward dynamics for temporal modeling.
\newblock In {\em 2023 IEEE International Conference on Data Mining (ICDM)}, pages 588--597. IEEE, 2023.

\bibitem{thompson1994modeling}
Michael~L Thompson and Mark~A Kramer.
\newblock Modeling chemical processes using prior knowledge and neural networks.
\newblock {\em AIChE Journal}, 40(8):1328--1340, 1994.

\bibitem{tran2021factorized}
Alasdair Tran, Alexander Mathews, Lexing Xie, and Cheng~Soon Ong.
\newblock Factorized fourier neural operators.
\newblock {\em arXiv preprint arXiv:2111.13802}, 2021.

\bibitem{tsai2021calibration}
Wen-Ping Tsai, Dapeng Feng, Ming Pan, Hylke Beck, Kathryn Lawson, Yuan Yang, Jiangtao Liu, and Chaopeng Shen.
\newblock From calibration to parameter learning: Harnessing the scaling effects of big data in geoscientific modeling.
\newblock {\em Nature communications}, 12(1):5988, 2021.

\bibitem{von2019informed}
Laura von Rueden, Sebastian Mayer, Katharina Beckh, Bogdan Georgiev, Sven Giesselbach, Raoul Heese, Birgit Kirsch, Julius Pfrommer, Annika Pick, Rajkumar Ramamurthy, et~al.
\newblock Informed machine learning--a taxonomy and survey of integrating knowledge into learning systems.
\newblock {\em arXiv preprint arXiv:1903.12394}, 2019.

\bibitem{wan2023molecules}
Yue Wan, Jialu Wu, Tingjun Hou, Chang-Yu Hsieh, and Xiaowei Jia.
\newblock From molecules to scaffolds to functional groups: building context-dependent molecular representation via multi-channel learning.
\newblock {\em arXiv preprint arXiv:2311.02798}, 2023.

\bibitem{wang2021deep}
Fang Wang, Di~Tian, Lisa Lowe, Latif Kalin, and John Lehrter.
\newblock Deep learning for daily precipitation and temperature downscaling.
\newblock {\em Water Resources Research}, 57(4):e2020WR029308, 2021.

\bibitem{wang2023scientific}
Hanchen Wang, Tianfan Fu, Yuanqi Du, Wenhao Gao, Kexin Huang, Ziming Liu, Payal Chandak, Shengchao Liu, Peter Van~Katwyk, Andreea Deac, et~al.
\newblock Scientific discovery in the age of artificial intelligence.
\newblock {\em Nature}, 620(7972):47--60, 2023.

\bibitem{wang2020incorporating}
Rui Wang, Robin Walters, and Rose Yu.
\newblock Incorporating symmetry into deep dynamics models for improved generalization.
\newblock {\em arXiv preprint arXiv:2002.03061}, 2020.

\bibitem{wang2023expert}
Sifan Wang, Shyam Sankaran, Hanwen Wang, and Paris Perdikaris.
\newblock An expert's guide to training physics-informed neural networks.
\newblock {\em arXiv preprint arXiv:2308.08468}, 2023.

\bibitem{wang2021understanding}
Sifan Wang, Yujun Teng, and Paris Perdikaris.
\newblock Understanding and mitigating gradient flow pathologies in physics-informed neural networks.
\newblock {\em SIAM Journal on Scientific Computing}, 43(5):A3055--A3081, 2021.

\bibitem{wang2021learning}
Sifan Wang, Hanwen Wang, and Paris Perdikaris.
\newblock Learning the solution operator of parametric partial differential equations with physics-informed deeponets.
\newblock {\em Science advances}, 7(40):eabi8605, 2021.

\bibitem{wang2022improved}
Sifan Wang, Hanwen Wang, and Paris Perdikaris.
\newblock Improved architectures and training algorithms for deep operator networks.
\newblock {\em Journal of Scientific Computing}, 92(2):35, 2022.

\bibitem{wang2021generative}
Zhengwei Wang, Qi~She, and Tomas~E Ward.
\newblock Generative adversarial networks in computer vision: A survey and taxonomy.
\newblock {\em ACM Computing Surveys (CSUR)}, 54(2):1--38, 2021.

\bibitem{wang2023high}
Zhihao Wang, Yiqun Xie, Xiaowei Jia, Lei Ma, and George Hurtt.
\newblock High-fidelity deep approximation of ecosystem simulation over long-term at large scale.
\newblock In {\em Proceedings of the 31st ACM International Conference on Advances in Geographic Information Systems}, pages 1--10, 2023.

\bibitem{wang2024simfair}
Zhihao Wang, Yiqun Xie, Zhili Li, Xiaowei Jia, Zhe Jiang, Aolin Jia, and Shuo Xu.
\newblock Simfair: Physics-guided fairness-aware learning with simulation models.
\newblock {\em arXiv preprint arXiv:2401.15270}, 2024.

\bibitem{wei2022mapping}
Zhihao Wei, Kebin Jia, Xiaowei Jia, Pengyu Liu, Ying Ma, Ting Chen, and Guilian Feng.
\newblock Mapping large-scale plateau forest in sanjiangyuan using high-resolution satellite imagery and few-shot learning.
\newblock {\em Remote Sensing}, 14(2):388, 2022.

\bibitem{wei2021large}
Zhihao Wei, Kebin Jia, Pengyu Liu, Xiaowei Jia, Yiqun Xie, and Zhe Jiang.
\newblock Large-scale river mapping using contrastive learning and multi-source satellite imagery.
\newblock {\em Remote Sensing}, 13(15):2893, 2021.

\bibitem{wijmans1995solution}
Johannes~G Wijmans and Richard~W Baker.
\newblock The solution-diffusion model: a review.
\newblock {\em Journal of membrane science}, 107(1-2):1--21, 1995.

\bibitem{willard2022integrating}
Jared Willard, Xiaowei Jia, Shaoming Xu, Michael Steinbach, and Vipin Kumar.
\newblock Integrating scientific knowledge with machine learning for engineering and environmental systems.
\newblock {\em ACM Computing Surveys}, 55(4):1--37, 2022.

\bibitem{willard2021predicting}
Jared~D Willard, Jordan~S Read, Alison~P Appling, Samantha~K Oliver, Xiaowei Jia, and Vipin Kumar.
\newblock Predicting water temperature dynamics of unmonitored lakes with meta-transfer learning.
\newblock {\em Water Resources Research}, 57(7):e2021WR029579, 2021.

\bibitem{wu2021ontology}
Jiantao Wu, Fabrizio Orlandi, Declan O'Sullivan, and Soumyabrata Dev.
\newblock An ontology model for climatic data analysis.
\newblock In {\em 2021 IEEE International Geoscience and Remote Sensing Symposium IGARSS}, pages 5739--5742. IEEE, 2021.

\bibitem{wu2022linkclimate}
Jiantao Wu, Fabrizio Orlandi, Declan O’Sullivan, and Soumyabrata Dev.
\newblock Linkclimate: An interoperable knowledge graph platform for climate data.
\newblock {\em Computers \& Geosciences}, 169:105215, 2022.

\bibitem{wu2023improving}
Jiantao Wu, Jarrett Pierse, Fabrizio Orlandi, Declan O'Sullivan, and Soumyabrata Dev.
\newblock Improving tourism analytics from climate data using knowledge graphs.
\newblock {\em IEEE Journal of Selected Topics in Applied Earth Observations and Remote Sensing}, 16:2402--2412, 2023.

\bibitem{wu2020enforcing}
Jin-Long Wu, Karthik Kashinath, Adrian Albert, Dragos Chirila, Heng Xiao, et~al.
\newblock Enforcing statistical constraints in generative adversarial networks for modeling chaotic dynamical systems.
\newblock {\em Journal of Computational Physics}, 406:109209, 2020.

\bibitem{xie2022statistically}
Yiqun Xie, Shashi Shekhar, and Yan Li.
\newblock Statistically-robust clustering techniques for mapping spatial hotspots: A survey.
\newblock {\em ACM Computing Surveys (CSUR)}, 55(2):1--38, 2022.

\bibitem{xie2023geo}
Yiqun Xie, Zhaonan Wang, Gengchen Mai, Yanhua Li, Xiaowei Jia, Song Gao, and Shaowen Wang.
\newblock Geo-foundation models: Reality, gaps and opportunities.
\newblock In {\em Proceedings of the 31st ACM International Conference on Advances in Geographic Information Systems}, pages 1--4, 2023.

\bibitem{xu2024knowledge}
Bo~Xu, Haofei Yu, Zongbo Shi, Jinxing Liu, Yuting Wei, Zhongcheng Zhang, Yanqi Huangfu, Han Xu, Yue Li, Linlin Zhang, et~al.
\newblock Knowledge-guided machine learning reveals pivotal drivers for gas-to-particle conversion of atmospheric nitrate.
\newblock {\em Environmental Science and Ecotechnology}, 19:100333, 2024.

\bibitem{xu2023physics}
Qingsong Xu, Yilei Shi, Jonathan Bamber, Ye~Tuo, Ralf Ludwig, and Xiao~Xiang Zhu.
\newblock Physics-aware machine learning revolutionizes scientific paradigm for machine learning and process-based hydrology.
\newblock {\em arXiv preprint arXiv:2310.05227}, 2023.

\bibitem{xu2023deep}
Yuanhao Xu, Kairong Lin, Caihong Hu, Shuli Wang, Qiang Wu, Li~Zhang, and Guang Ran.
\newblock Deep transfer learning based on transformer for flood forecasting in data-sparse basins.
\newblock {\em Journal of Hydrology}, 625:129956, 2023.

\bibitem{yan2023adaptive}
Jiexi Yan, Lei Luo, Cheng Deng, and Heng Huang.
\newblock Adaptive hierarchical similarity metric learning with noisy labels.
\newblock {\em IEEE Transactions on Image Processing}, 32:1245--1256, 2023.

\bibitem{yang2023flexible}
Qi~Yang, Licheng Liu, Junxiong Zhou, Rahul Ghosh, Bin Peng, Kaiyu Guan, Jinyun Tang, Wang Zhou, Vipin Kumar, and Zhenong Jin.
\newblock A flexible and efficient knowledge-guided machine learning data assimilation (kgml-da) framework for agroecosystem prediction in the us midwest.
\newblock {\em Remote Sensing of Environment}, 299:113880, 2023.

\bibitem{yang2023fourier}
Qidong Yang, Alex Hernandez-Garcia, Paula Harder, Venkatesh Ramesh, Prasanna Sattegeri, Daniela Szwarcman, Campbell~D Watson, and David Rolnick.
\newblock Fourier neural operators for arbitrary resolution climate data downscaling.
\newblock {\em arXiv preprint arXiv:2305.14452}, 2023.

\bibitem{yang2019enforcing}
Zeng Yang, Jin-Long Wu, and Heng Xiao.
\newblock Enforcing deterministic constraints on generative adversarial networks for emulating physical systems.
\newblock {\em arXiv preprint arXiv:1911.06671}, 2019.

\bibitem{yeganeh2022machine}
Abbas Yeganeh-Bakhtiary, Hossein EyvazOghli, Naser Shabakhty, Bahareh Kamranzad, and Soroush Abolfathi.
\newblock Machine learning as a downscaling approach for prediction of wind characteristics under future climate change scenarios.
\newblock {\em Complexity}, 2022, 2022.

\bibitem{zahura2022predicting}
Faria~T Zahura and Jonathan~L Goodall.
\newblock Predicting combined tidal and pluvial flood inundation using a machine learning surrogate model.
\newblock {\em Journal of Hydrology: Regional Studies}, 41:101087, 2022.

\bibitem{zhang2022prediction}
Di~Zhang, Dongsheng Wang, Qidong Peng, Junqiang Lin, Tiantian Jin, Tiantian Yang, Soroosh Sorooshian, and Yi~Liu.
\newblock Prediction of the outflow temperature of large-scale hydropower using theory-guided machine learning surrogate models of a high-fidelity hydrodynamics model.
\newblock {\em Journal of Hydrology}, 606:127427, 2022.

\bibitem{zhang2023artificial}
Xuan Zhang, Limei Wang, Jacob Helwig, Youzhi Luo, Cong Fu, Yaochen Xie, Meng Liu, Yuchao Lin, Zhao Xu, Keqiang Yan, et~al.
\newblock Artificial intelligence for science in quantum, atomistic, and continuum systems.
\newblock {\em arXiv preprint arXiv:2307.08423}, 2023.

\bibitem{zhao2022incremental}
Jiawei Zhao, Robert~Joseph George, Yifei Zhang, Zongyi Li, and Anima Anandkumar.
\newblock Incremental fourier neural operator.
\newblock {\em arXiv preprint arXiv:2211.15188}, 2022.

\bibitem{zhao2023survey}
Wayne~Xin Zhao, Kun Zhou, Junyi Li, Tianyi Tang, Xiaolei Wang, Yupeng Hou, Yingqian Min, Beichen Zhang, Junjie Zhang, Zican Dong, et~al.
\newblock A survey of large language models.
\newblock {\em arXiv preprint arXiv:2303.18223}, 2023.

\bibitem{zhu2022building}
Qing Zhu, Fa~Li, William~J Riley, Li~Xu, Lei Zhao, Kunxiaojia Yuan, Huayi Wu, Jianya Gong, and James Randerson.
\newblock Building a machine learning surrogate model for wildfire activities within a global earth system model.
\newblock {\em Geoscientific Model Development}, 15(5):1899--1911, 2022.

\bibitem{zhuang2020comprehensive}
Fuzhen Zhuang, Zhiyuan Qi, Keyu Duan, Dongbo Xi, Yongchun Zhu, Hengshu Zhu, Hui Xiong, and Qing He.
\newblock A comprehensive survey on transfer learning.
\newblock {\em Proceedings of the IEEE}, 109(1):43--76, 2020.

\end{thebibliography}

\end{document}